%% file: main.tex
\PassOptionsToPackage{table,xcdraw,dvipsnames}{xcolor}

\documentclass[10pt,twocolumn,letterpaper]{article}

\usepackage[pagenumbers]{cvpr}              
\input{preamble}
\input{command}

\title{Imagine Before Concentration: Diffusion-Guided Registers \\ Enhance Partially Relevant Video Retrieval}

\author{
Jun Li$^{1,2,3}$, \ 
Xuhang Lou$^{1}$, \ 
Jinpeng Wang$^{1}\thanks{Corresponding author.}$\;, \
Yuting Wang, \\
Yaowei Wang$^{1,3}$, \ 
Shu-Tao Xia$^{2,3}$, \ 
Bin Chen$^{1,3}$ \\
{\small $^1$Harbin Institute of Technology, Shenzhen} \\
{\small $^2$Tsinghua Shenzhen International Graduate School, Tsinghua University} \\
{\small $^3$Peng Cheng Laboratory} \\
{\small \texttt{220110924@stu.hit.edu.cn}\quad\Letter\ \texttt{wangjp26@gmail.com}}
}

\begin{document}
\maketitle
\input{sections/Abstract}    
\input{sections/Introduction}

\input{sections/RelatedWorks}

\input{sections/Method}
\input{sections/Experiments}
\input{sections/Conclusions}
{
    \small
    \bibliographystyle{ieeenat_fullname}
    \bibliography{main}
}

\clearpage
\maketitlesupplementary

\thispagestyle{empty}
\appendix

\input{appendix_sections/Method}

\end{document}

%% file: preamble.tex
\usepackage{graphicx}
\usepackage{amsmath}
\usepackage{amssymb}
\usepackage{booktabs}

\definecolor{cvprblue}{rgb}{0.21,0.49,0.74}
\usepackage[pagebackref,breaklinks,colorlinks,allcolors=cvprblue]{hyperref}

\usepackage{marvosym}
\usepackage{url}            
\usepackage{amsfonts}       
\usepackage{nicefrac}       
\usepackage{microtype}      

\usepackage{makecell}
\usepackage{amsthm}
\usepackage{bm}
\usepackage[table,xcdraw]{xcolor}
\usepackage{multicol}
\usepackage{colortbl}
\usepackage{multirow}
\usepackage{enumitem}
\usepackage{color}

\usepackage{algorithm}
\usepackage{listings}
\usepackage{natbib}
\usepackage{algorithmic}

\usepackage{etoolbox}
\makeatletter
\AfterEndEnvironment{algorithm}{\let\@algcomment\relax}
\AtEndEnvironment{algorithm}{\kern2pt\hrule\relax\vskip3pt\@algcomment}
\let\@algcomment\relax
\newcommand\algcomment[1]{\def\@algcomment{\footnotesize#1}}
\renewcommand\fs@ruled{\def\@fs@cfont{\bfseries}\let\@fs@capt\floatc@ruled
  \def\@fs@pre{\hrule height.8pt depth0pt \kern2pt}%
  \def\@fs@post{}%
  \def\@fs@mid{\kern2pt\hrule\kern2pt}%
  \let\@fs@iftopcapt\iftrue}
\makeatother

\usepackage[table]{xcolor}

\definecolor{ourscolor}{HTML}{E3EFDB}

\definecolor{basegray}{HTML}{F5F5F5}  %

%% file: command.tex
\definecolor{ForestGreen}{RGB}{34,139,34}

\renewcommand{\paragraph}[1]{\medskip\noindent\textbf{#1.~}}

\newcommand{\bmc}{\bm{c}}

\newcommand{\bmf}{\bm{f}}

\newcommand{\bmq}{\bm{q}}
\newcommand{\bmr}{\bm{r}}

\newcommand{\bmx}{\bm{x}}

\newcommand{\bmV}{\bm{V}}

\definecolor{darkred}{RGB}{167,21,0}   %
\definecolor{darkblue}{RGB}{0,81,127}    
\newcommand{\modelname}{\textbf{\textcolor{darkred}{Dream}\textcolor{darkblue}{PRVR}}}

%% file: sections/Abstract.tex
\begin{abstract}
Partially Relevant Video Retrieval (PRVR) aims to retrieve untrimmed videos based on text queries that describe only partial events. Existing methods suffer from incomplete global contextual perception, struggling with  query ambiguity and local noise induced by spurious responses.
To address these issues, we propose \modelname{}, which adopts a coarse-to-fine representation learning paradigm. The model first generates global contextual semantic registers as coarse-grained highlights spanning the entire video and then concentrates on  fine-grained similarity optimization for precise cross-modal matching.
Concretely, these registers are generated by  initializing from the video-centric distribution produced by a probabilistic variational sampler and then iteratively refined via a text-supervised truncated diffusion model. During this process, textual semantic structure learning constructs a well-formed textual latent space, enhancing the reliability of global perception. The registers are then adaptively fused  with video tokens through register-augmented Gaussian attention blocks, enabling context-aware feature learning.
Extensive experiments show that \modelname{} outperforms state-of-the-art methods. Code is released at 
\href{https://github.com/lijun2005/CVPR26-DreamPRVR}{https://github.com/lijun2005/CVPR26-DreamPRVR}.
\end{abstract}

%% file: sections/Introduction.tex
\section{Introduction}
\label{sec:intro}
\begin{figure*}[t]
    \centering
    \includegraphics[width=\textwidth]{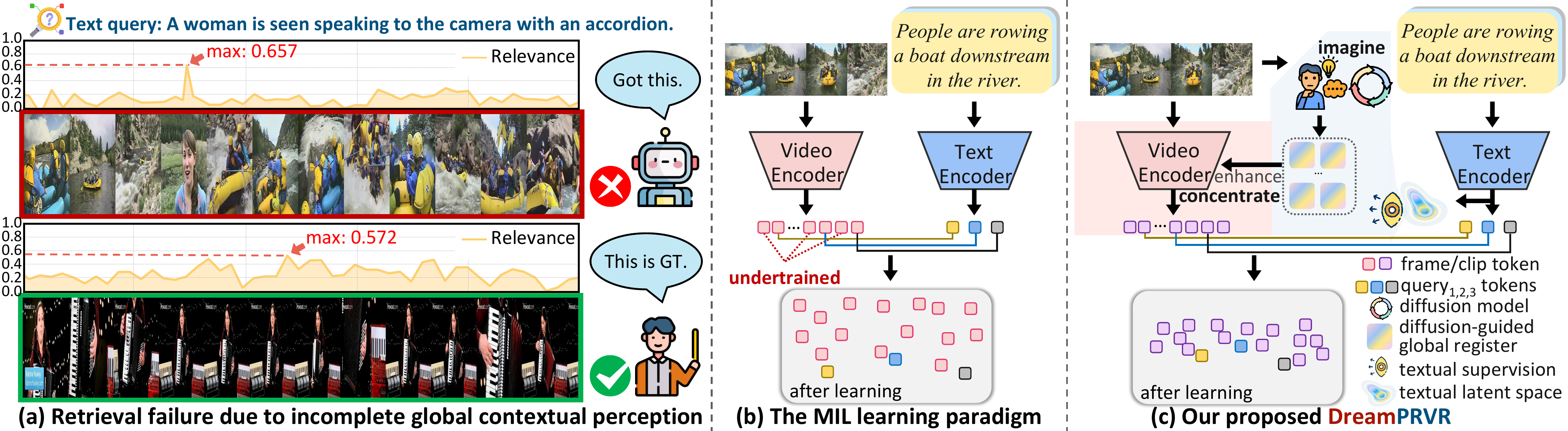}
    \vspace{-1.5em}
    \caption{(a)
    Limited contextual awareness causes \emph{spurious local spikes} on a globally irrelevant ``people boating'' video, incorrectly outscoring the ground-truth ``demonstrating the accordion through the camera'' video which exhibits more overall relevance.
    (b) MIL considers only the closest pair, leading to \emph{sparse supervision} where others remain undertrained.
    (c) DreamPRVR first \emph{imagines}  global registers through text-supervised diffusion, then \emph{concentrates} on fine-grained learning,
    thereby realizing \emph{coarse-to-fine} cross-modal alignment and jointly optimizing all video tokens to form a coherent embedding space.
    }
    \label{fig:Intro}
    \vspace{-.5em}
\end{figure*}
Text-to-Video Retrieval (T2VR) \cite{Video-ColBERT,BLiM,UMIVR,PIG,dong2022reading}  facilitates access to large-scale video collections. While most  T2VR methods assume a fully relevant setting, where short trimmed videos perfectly match the query,
this premise may contrast with real-world applications, where untrimmed videos are common and queries often describe only a partial clip. To bridge this gap,  \emph{Partially Relevant Video Retrieval} (PRVR) \cite{ms-sl} was introduced, which aims to  retrieve untrimmed videos based on  given partial relevant queries.

A core challenge in PRVR lies in \emph{query ambiguity} \cite{RAL,ARL}, where a general query may match its ground-truth clip while inadvertently  aligns with local clips from other videos. 
Such ambiguous associations inject noise into  precise cross-modal matching, degrading retrieval accuracy.
We attribute this to  \emph{incomplete global contextual perception}  in the partially relevant learning setting, where queries may be easily misled by  globally irrelevant videos  with coincidentally similar events, producing local spiky activation responses and retrieval failures, as shown in \cref{fig:Intro}(a). Moreover,  \cref{fig:Intro}(b) shows that the widely used multiple instance learning paradigm (MIL) in PRVR  exacerbates this issue by rewarding only the best-matching clip, leaving others undertrained and lacking sufficient contextual grounding to resolve ambiguity.

Despite growing attention to global context, most PRVR methods \cite{ms-sl,GMMFORMER,PEAN} lack explicit modeling. HLFormer \cite{HLFormer} models semantic entailment and RAL \cite{RAL} captures global uncertainty, yet both treat global context as training-only regularization, leaving video embeddings unrefined at inference. DL-DKD \cite{DL-DKD} introduces CLIP-based global knowledge, but its teacher model remains temporally constrained. 
Motivated by register tokens \cite{vit-reg,mamba-reg,smallregisters}, we also introduce global registers to store holistic video semantics, which interact with all tokens to enhance local representations, mitigate MIL under-training and suppress noisy local spike activations.

However, extracting reliable global registers remains non-trivial due to redundancy and noise in untrimmed videos.
Simple pooling or one-step mapping methods may fail to disentangle trustworthy semantics. 
To address this, we propose \modelname{}, performing \emph{coarse-grained contextual imagination before concentrating on fine-grained representation learning}. As depicted in \cref{fig:Intro}(c), it first generates global registers through a text-supervised diffusion process that iteratively denoises and refines video semantics, yielding reliable holistic context to enhance representation learning. 

Executing the generation  entails two core challenges: (\textbf{i})  obtaining reliable textual supervision to guide semantic generation and (\textbf{ii})  decoupling trustworthy global registers from noisy untrimmed videos. 
For (i), existing query diversity loss \cite{GMMFORMER,GMMFormerV2,HLFormer} blindly separates all queries, ignoring intra-video  correlations. We address with a Query Similarity Preservation (QSP) Loss , treating queries from the same video as complementary positive views of its global semantics. Together with the diversity loss, QSP jointly enhances intra-video compactness and inter-video separability, yielding a well-structured latent space.
To further explicitly model this space under textual uncertainty, we introduce a Textual Perturbation Sampler (TPS), which approximates the latent space by sampling within controllable perturbations, providing  semantically aligned supervision for register generation.
For (ii),  a truncated diffusion model is designed to facilitate generation. Rather than initializing from random noise, a Probabilistic Variational Sampler (PVS) constructs a learnable probabilistic latent space to generate video-centric distribution as a semantically grounded generation starting point. Guided by TPS, the Diffusion Register Estimator (DRE) then performs iterative refinement, progressively denoising PVS-initialized embeddings into pure and holistic registers encoding comprehensive contextual semantics.
Finally, the Register-augmented Gaussian Attention Blocks (RAB) adaptively fuse the refined registers with video tokens via asymmetric contextual attention masks,  enhancing fine-grained representation learning in the end.

Empirical results on ActivityNet Captions \cite{krishna2017dense}, Charades-STA \cite{gao2017tall} and TVR \cite{lei2020tvr} validate the state-of-the-art performance of \modelname{}.
Unlike large-scale diffusion models \cite{Imagen,dit}, our model is lightweight, requiring only a few registers and timesteps to achieve notable  gains, highlighting high efficiency. Extensive ablations and visualizations further confirm that the registers progressively acquire cleaner semantics through iterative refinement,
while  textual semantic structure learning enforces a well-organized latent space.

To summarize, we make the following contributions.
\begin{itemize}
    \item[$\bullet$]  We propose \modelname{}, a \emph{contextual imagination} framework for PRVR,   which first generates registers to capture coarse-grained holistic video semantics and then concentrates on  fine-grained representation learning,  achieving hierarchical and progressive cross-modal alignment.
    \item[$\bullet$] 
    We supervise register generation with a structured textual space constructed via textual semantic structure learning, estimate registers through a truncated diffusion model initialized from the video-centric distribution and employ register-augmented Gaussian attention for adaptive fusion.
    \item[$\bullet$]  Extensive experiments validate our model's superiority, with analyses showing effectiveness of global registers and acceptable efficiency of  the imagination process.
\end{itemize}

%% file: sections/RelatedWorks.tex
\section{Related Works}
\label{sec:related_works}
\noindent \textbf{Partially Relevant Video Retrieval} \quad 
Text-based video retrieval \cite{lin2024multigranularity} is a core area of information retrieval \cite{shi2025multi,tang2025reason,wang2025efficient,lian2025autossvh}. Text-to-Video Retrieval (T2VR) \cite{Video-ColBERT, PIG, BLiM, UMIVR, tian2024holistic} retrieves fully relevant pre-trimmed videos. Video Corpus Moment Retrieval (VCMR) \cite{song2021spatial, lei2020tvr,jsg} localizes specific temporal moments across large video corpora. Partially Relevant Video Retrieval (PRVR) \cite{MS-SL++, uem, amdnet, ProtoPRVR, bgmnet, ARL,msc-prvr,sdm-prvr}, introduced by \citet{ms-sl} extends retrieval to untrimmed videos where only partial segments match the query.
Existing PRVR methods primarily emphasize clip-level modeling. MS-SL \cite{ms-sl} constructs dense clip embeddings via a sliding-window scheme, while ProtoPRVR \cite{ProtoPRVR} accelerates retrieval by employing representative prototypes. GMMFormer \cite{GMMFORMER} introduces  implicit clip modeling based on Gaussian attention and HLFormer \cite{HLFormer} leverages hyperbolic space to capture hierarchical video structures. For learning objectives, DL-DKD \cite{DL-DKD} distills dynamic CLIP \cite{CLIP} knowledge, ARL \cite{ARL} formulates query ambiguity with dedicated optimization goals and RAL \cite{RAL} advances probabilistic query–video embedding learning.
Despite these efforts, global contextual semantics remain underexplored, typically ignored or introduced as training-only losses, leading to limited global awareness.
To address this gap, we propose diffusion-guided registers that explicitly encode holistic video context and provide global cues during both training and retrieval.

\noindent \textbf{Registers in Vision Models} \quad 
Registers, initially introduced by \citet{vit-reg}, are learnable tokens appended to Vision Transformer \cite{ViT} inputs to mitigate high-norm token issues
while retaining global image information. They have demonstrated consistent advantages across diverse vision tasks. For instance, Mamba-Reg \cite{mamba-reg} integrates registers into Vision Mamba to enhance scalability. FALCON \cite{FALCON} employs visual registers in high-resolution MLLMs to alleviate visual redundancy and semantic fragmentation. RegQAV \cite{RegQAV} incorporates them into audio-visual foundation models for improved forgery localization. Unlike prior works, \modelname{} exploits registers to provide reliable contextual cues for retrieval, seamlessly integrating the generative capability of diffusion models into the retrieval process.

\noindent \textbf{Diffusion Models} \quad
Diffusion models \cite{DDPM, DDIM} have emerged as powerful generative frameworks, extended from image synthesis \cite{LDM, Imagen, DALL-E2} to multi-modal domains such as video generation \cite{sun2025ar, qi2025mask}, semantic segmentation \cite{wang2025vidseg, rahman2023ambiguous}, and music synthesis \cite{schneider2024mousai, chowdhury2024melfusion}.
Recent advances further adapt diffusion paradigms for retrieval: DiffDis \cite{DiffDis} formulates retrieval as generative diffusion of text embeddings; DiffusionRet \cite{diffusionret} models the joint distribution of queries and candidates; MomentDiff \cite{momentdiff} iteratively refines random spans into moments; and DITS \cite{DITS} achieves text–video alignment via direct diffusion generation. Distinct from these approaches, we incorporate registers to furnish reliable global context within the diffusion-based refinement process, enhancing fine-grained feature learning while suppressing local noise.

%% file: sections/Method.tex
\section{Method}
\label{sec:method}
\begin{figure*}[t]
    \centering
    \includegraphics[width=\textwidth]{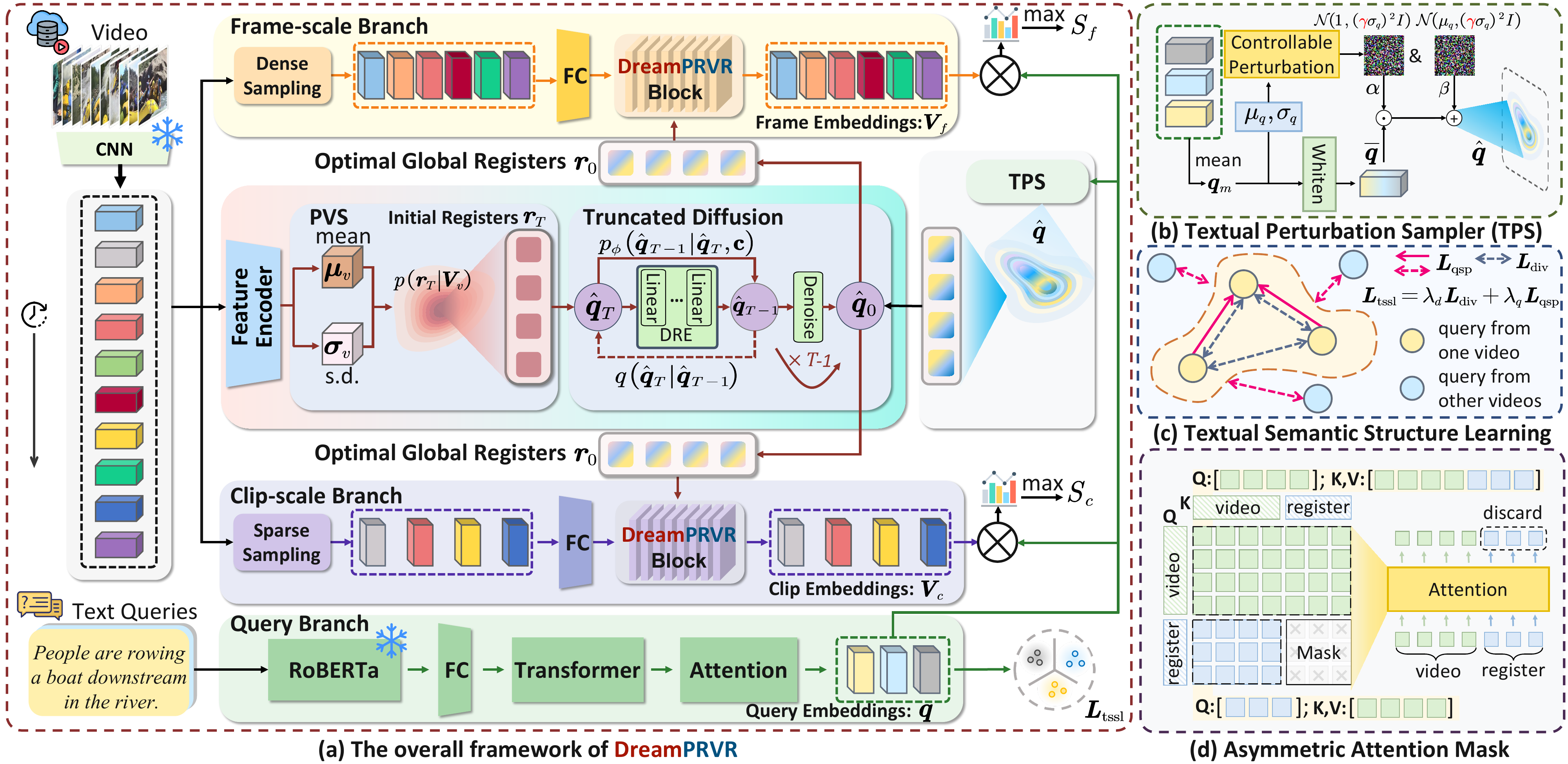}
    \vspace{-2em}
    \caption{Overview of \modelname{}. 
    (\textbf{a}) The query branch produces embedding $\bmq$ and samples $\hat{\bmq}$ via TPS to supervise register generation. Video embeddings from a pre-trained model are first processed by a lightweight feature encoder and a probabilistic variational sampler (PVS) to produce the initial register ${\bmr}_{T}$, which is iteratively denoised via a truncated diffusion model to yield optimal registers ${\bmr}_{0}$.  ${\bmr}_{0}$ subsequently enhance frame- and clip-level representation learning, getting frame embeddings $\bmV_f$ and clip embeddings $\bmV_c$.
     $\bmq$ learns a latent semantic structure through $L_{\text{tssl}}$ and computes similarity scores $S_f$ and $S_c$.
    (\textbf{b}) Textual Perturbation Sampler (TPS) models query uncertainty via controllable perturbations and samples $\hat{\bmq}$ without trainable parameters.
    (\textbf{c}) Textual Semantic Structure Learning $L_{\text{tssl}}$ employs $L_{\text{div}}$ to diversify queries and $L_{\text{qsp}}$ to align queries from the same video while contrasting across videos. (\textbf{d}) The asymmetric attention mask defines two  cross-attention patterns, enabling full interactions for video tokens while constraining registers to video-only attention.
    }
    \vspace{-1em}
    \label{fig:arc}
\end{figure*}

\subsection{Problem Formulation and Overview}
Partially Relevant Video Retrieval (PRVR) seeks to retrieve videos $V \in \mathcal{V}$ that contain moments relevant to a text query $Q$, where each video comprises multiple moment-description pairs and each query targets a specific moment. In this paper, we propose \modelname{}, which first generates diffusion-guided global registers $r$ to capture holistic contextual semantics and then enhances fine-grained  video  representations to suppress spurious local responses  for better retrieval.

\noindent \textbf{Probabilistic Pipeline Formulation}  \quad Given a  text-video pair $(Q, V)$, the registers $r$ are inferred solely from $V$. The overall model pipeline is formulated as:
\vspace{-.8em}
\begin{equation}
    p_{\theta,\phi}(Q|V) = \int \underbrace{p_\theta(Q|V,r)}_{\text{concentration}} \cdot 
    \underbrace{p_\phi(r|V)}_{\text{imagination}}  dr,
    \label{eq:all-register}
    \vspace{-.8em}
\end{equation}
where $\theta$ and $\phi$ denote the parameters of the alignment model, including the video and text encoders, and the register generator (\textit{i.e.}, diffusion model), respectively.

\noindent\textbf{Variational Inference for Register Generation} \quad 
Inspired by VAE \cite{vae}, we employ variational inference to establish a principled optimization framework. Unlike untrimmed and noisy videos, textual queries are concise and directly capture the video's holistic semantics, which we use to supervise register generation.  Based on this, we introduce a variational posterior $q_\varphi(r| Q_a)$ to approximate the true posterior $p(r | Q_a)$, where $Q_a$ denotes all textual queries associated with a video, and $\varphi$ are the parameters of the network.
The Evidence Lower Bound (ELBO) of \cref{eq:all-register} is then given by:
\begin{equation}
    \begin{aligned}
        \log p_{\theta, \phi}(Q | V) &\geq \mathbb{E}_{q_\varphi(r | Q_a)} \left[ \log p_\theta(Q | V, r) \right] \\
        &\quad - \mathbb{KL} \left[ q_\varphi(r | Q_a) \, \| \, p_\phi(r | V) \right].
    \end{aligned}  
    \label{eq:elbo}
\end{equation}
The derivation of \cref{eq:elbo} is provided in Appendix.
Following VAE \cite{vae}, we maximize the ELBO in \cref{eq:elbo} to optimize \cref{eq:all-register}. The optimization function is defined as:
\begin{equation}
\begin{aligned}
    L_\text{DreamPRVR} = &- \mathbb{E}_{q_\varphi(r | Q_a)} \left[ \log p_\theta(Q | V, r) \right]\\ &+ \mathbb{KL} \left[ q_\varphi(r | Q_a) \, \| \, p_\phi(r | V) \right].
\end{aligned}
\label{eq:optim}
\end{equation}
This comprises two objectives: (i) minimizing the KL divergence to enforce the registers’ encoding of video semantics conveyed by textual queries, implying that the entire register generation process is guided by textual supervision and (ii) maximizing the likelihood term to enhance video representation learning with registers for cross-modal retrieval. In practice, we optimize the model by making $p_\phi(r|V)$  progressively approximate  $q_\varphi(r|Q_a)$ and sampling  $r$ accordingly.

\noindent \textbf{Framework Overview} \quad Following the above principle, \modelname{} is carefully designed with four core components: textual semantic structure learning, global register generation, register-augmented video representation and similarity computation, as illustrated in \cref{fig:arc}.

\subsection{Textual Semantic Structure Learning}
Given a text query of $N_q$ words, word-level features are extracted via a pre-trained RoBERTa \cite{liu2019roberta} and projected to a lower-dimensional space through a fully connected layer. A standard Transformer \cite{vaswani2017attention} encoder produces query representations $Q \in \mathbb{R}^{N_q \times d}$, which are aggregated into the final embedding $\bmq \in \mathbb{R}^d$ using the attention mechanism from MS-SL \cite{ms-sl}. All  embeddings are used to construct a structured  latent space via a combination of query similarity preservation and diversity losses, explicitly modeled by the Textual Perturbation Sampler (TPS) to supervise register generation.

\noindent \textbf{Query Similarity Preservation} \quad Existing methods \cite{GMMFORMER, GMMFormerV2, HLFormer} employ a query diversity loss $L_{\text{div}}$ that blindly separates queries, 
ignoring the  shared global semantic theme of the video.
Therefore, as shown in \cref{fig:arc} (c), we introduce a Query Similarity Preservation (QSP) Loss, aligning queries from the same video as complementary positives while contrasting those from different videos. For the $i$-th query embedding $\bmq_i$, the loss is:
\begin{equation}
   \!\!\!\!\! L_{\text{qsp}} \!\!=\! -\frac{1}{|V_{q}(i)|} \! \sum_{j \in V_{q}(i)} \!\!\!\log \frac{\exp(\text{sim}(\bmq_i, \bmq_j) / \tau)}{\sum_{k \in \Omega} \exp(\text{sim}(\bmq_i, \bmq_k) / \tau)},
    \label{eq:single-qsp}
\end{equation}
where $V_{q}(i)$ denotes the indices of queries from the same video as $\bmq_i$ with cardinality $|V_{q}(i)|$, $\Omega$ represents all query indices and $\tau$ is a temperature coefficient. Finally, the overall objective for textual semantic structure learning is:
\begin{equation}
    L_{\text{tssl}} = \lambda_{d} L_{\text{div}} +  \lambda_{q} L_{\text{qsp}},
    \label{eq:tssl-loss}
\end{equation}
where  $\lambda_{q}$ and $\lambda_{d}$ are weights. While the diversity loss enriches semantics by separating queries, QSP preserves intra-video query similarity and enhances inter-video discriminability, forming a well-structured  latent space.

\noindent \textbf{Textual Perturbation Sampler} \quad
As shown in \cref{fig:arc} (b), to explicitly model the textual latent space, we first compute a global semantic representation $\bmq_m$ by averaging all query  embeddings of a video. Owing to the inherent uncertainty of queries \cite{ tang2025modeling, chen2024composed,duan2025fuzzy,tang2026heterogeneous}, deterministic point-wise modeling may fail to capture their variability. Thus, TPS approximates the latent space via controlled  perturbations, injecting noise into the whitened feature 
$\bar{\bmq} = (\bmq_m - \mu_q)/  \sigma_q$, as follows:
\begin{equation} 
\hat{\bmq} = \alpha \cdot \bar{\bmq} + \beta \quad \in \mathbb{R}^{d},
\label{eq:tps-sample}
\end{equation}
where $\alpha \sim \mathcal{N}(1, (\gamma\sigma_q)^2I)$, $\beta \sim \mathcal{N}(\mu_q, (\gamma\sigma_q)^2I)$,  $\mu_q$ and $\sigma_q$ are computed from ${q}_m$ and $\gamma$ denotes the perturbation scale. These features, with limited variation, capture uncertainty while adhering to the query distributions.

\subsection{Register Generation via Truncated Diffusion}
Based on \cref{eq:optim}, the model is tasked with generating global registers that capture the video semantics conveyed by textual queries. To this end, textual features ${\hat{\bmq}}$  explicitly sampled via TPS (\cref{eq:tps-sample}) serve as generation targets and supervise the process. 
Inspired by \cite{luo2022understanding}, we formulate the generation process over iterative timesteps $T$, as follows:
\begin{equation}
\setlength{\abovedisplayskip}{0pt}
\setlength{\belowdisplayskip}{0pt}
\begin{minipage}[c]{0.9\linewidth}
\begin{align*}
 &\log p({\hat{\bmq}})\\[-10pt] 
 \ge& \mathbb{E}_{q({\hat{\bmq}}_{1:T}|{\hat{\bmq}}_0)}\Big[\log \frac{p({\hat{\bmq}}_{0:T})}{q({\hat{\bmq}}_{1:T}|{\hat{\bmq}}_0)}\Big] \\[-4pt] 
 =& \mathbb{E}_{q({\hat{\bmq}}_1|{\hat{\bmq}}_0)}[\log p_{\phi}({\hat{\bmq}}_0|{\hat{\bmq}}_1)] \\[-4pt]  
 &- \mathbb{E}_{q({\hat{\bmq}}_{T-1}|{\hat{\bmq}}_0)}\Big[\mathbb{KL}(q({\hat{\bmq}}_T|{\hat{\bmq}}_{T-1}) \| p({\hat{\bmq}}_T))\Big] \\[-4pt]  
 &- \sum_{t=1}^{T-1} \mathbb{E}_{q({\hat{\bmq}}_{t-1},{\hat{\bmq}}_{t+1}|{\hat{\bmq}}_0)}\Big[\mathbb{KL}(q({\hat{\bmq}}_t|{\hat{\bmq}}_{t-1}) \| p_{\phi}({\hat{\bmq}}_t|{\hat{\bmq}}_{t+1}))\Big],
\end{align*}
\end{minipage}\hfill
\raisebox{0.5\height}{(\theequation)}   %
\label{eq:elbo-allgenerate}
\end{equation}
where $\phi$ denotes the generator.
Optimizing the complex objectives in \cref{eq:elbo-allgenerate} is  challenging. Fortunately, diffusion models (DMs) \cite{DDPM, DDIM} offer a powerful solution with strong generative capabilities. Here, in contrast to large-scale DMs, we design a lightweight  truncated diffusion model to generate registers, comprising the below two components.

\noindent \textbf{Probabilistic Variational Sampler} \quad In \cref{fig:arc}  (a), embeddings of an untrimmed video are first extracted via a pre-trained vision model, and then processed by a lightweight feature encoder composed of a linear layer and a standard Transformer block, producing $\bmV_{v} \in \mathbb{R}^{N_v \times d}$. Next, $\bmV_v$ is fed into PVS to define a probabilistic embedding space as a normal distribution $p(\bmr_T |\bmV_v)$ with mean vectors and diagonal covariance matrices in $\mathbb{R}^d$:
\begin{equation}
    p(\bmr_T|\bmV_v) \sim \mathcal{N} (\bm\mu_v, 
                    \bm\sigma^2_v I),   
\label{eq:distribution}
\end{equation}
where the mean $\bm\mu_v$ is computed by a Fully Connected (FC) layer followed by LayerNorm \cite{ba2016layer} and $l_2$ normalization, the s.d. $\bm\sigma_v$ by a separate FC layer without normalization following \cite{pcme, uatvr}. Then we sample $N_r$ instances from $p(\bmr_T|V_v)$ to obtain video-centric noise $\bmr_T$ via reparameterization \cite{repara}:
\begin{equation}
\bmr_T = \bm\sigma_v \cdot \eta + \bm\mu_v\in \mathbb{R}^{N_r \times d}, \quad \eta \sim \mathcal{N}(0, I).
\label{eq:sample}
\end{equation}
In fact, $\bmr_T$ represents the initial state of the global registers, with $N_r$ denoting their number. Rather than initializing from random noise, our model generates a video-centric distribution $\bmr_T$ as the starting point, enabling truncated generation.

Following \cite{HIB, cliff}, we compute a KL divergence $L_{\text{kl}}$ between $p(\bmr_T|V_v)$ and the  prior $\mathcal{N}(0,I)$ to enforce the Gaussian constraint required by the diffusion formulation \cite{DDPM}.
Thus, the overall objective of PVS is:
$ L_{\text{pvs}} = \lambda_{kl} L_{\text{kl}}$. 

\noindent \textbf{Diffusion Register Estimator} \quad As shown in \cref{fig:arc} (a), with video-centric starting point $\bmr_T$, we employ a diffusion process to estimate the target $\hat{\bm q}$, generating the optimal registers $\bmr_0$.  We design a lightweight MLP-based diffusion module $\epsilon_{\phi}(\cdot)$, with details  provided in the Appendix.

We treat the target $\hat{\bm q}$ as clean data $\hat{\bmq}_0$ and follow the DDPM \cite{DDPM} framework, which gradually injects Gaussian noise through the fixed forward process
$q({\hat{\bmq}}_t|{\hat{\bmq}}_{t-1})$ for $t = 1, \dots, T$, which can be simply expressed as:
\begin{equation}
q({\hat{\bmq}}_{t}|{\hat{\bmq}}_{0}) = \mathcal{N}({\hat{\bmq}}_t;\sqrt{\bar{\alpha}_T}{\hat{\bmq}}_0, (1-\bar{\alpha}_t)I),
    \label{eq:forward}
\end{equation}
where $\bar{\alpha}_t$ defines the noise schedule.  DRE aims to recover the clean data from $\bmr_T$ instead of random Gaussian noise, denoted $\hat{\bmq}_T$, via a learned reverse process conditioned on $\mathbf{c}$: 
\begin{equation}
p_{\phi}({\hat{\bmq}}_{0:T} | \mathbf{c}) = p({\hat{\bmq}}_T) \prod_{t=1}^{T} p_{\phi}({\hat{\bmq}}_{t-1} | {\hat{\bmq}}_t, \mathbf{c}),
\label{eq:reverse-all}
\end{equation}
with each step given by:
\begin{equation}
    {\hat{\bmq}}_{t-1} = \frac{1}{\sqrt{\alpha_t}} \left( {\hat{\bmq}}_t - \frac{1-\alpha_t}{\sqrt{1-\bar{\alpha}_t}} \boldsymbol{\epsilon}_{\phi}({\hat{\bmq}}_t, \mathbf{c}, t) \right) + \sigma_t \mathbf{z},
 \label{eq:reverse-eachstep}
\end{equation}
where $\sigma_t$ is a predefined parameter and $\mathbf{z} \sim \mathcal{N}(0, I)$, which can also be sampled from a PVS-defined video-centric noise space. The condition $\mathbf{c} \in \mathbb{R}^{N_r \times d}$ is obtained through a simple cross-attention between $\bmV_v$ and learnable parameters.
In the reverse process, the DRE $\boldsymbol{\epsilon}_{\phi}(\cdot)$ estimates the noise added to each intermediate noisy input from \cref{eq:forward}. Therefore, the objective of DRE is defined as:
\begin{equation}
L_{\text{dre}} = \mathbb{E}_{t, \hat{\bmq}_t, \boldsymbol{\epsilon}} \left[ \left\| \boldsymbol{\epsilon} - \boldsymbol{\epsilon}_{\phi}(\hat{\bmq}_t, t, \mathbf{c}) \right\|^2 \right],
\label{eq:dre-loss}
\end{equation}
where $\boldsymbol{\epsilon}$ denotes the added noise in the forward process \cref{eq:forward}. We  finally iteratively apply  \cref{eq:reverse-eachstep} for $T$ steps to produce the best global registers $\bmr_0$ that approximate $\hat{\bmq}_0$.
\subsection{Register-Augmented Video Representation}
Following prior works \cite{ms-sl,HLFormer}, we adopt a dual-branch architecture.
The frame-scale branch densely samples $M_f$ frames, projects them to dimension $d$ via an FC layer, and refines them through the \modelname{} block to obtain frame embeddings  $ \bm{V}_f = \{\bmf_i\}_{i=1}^{M_f} \in \mathbb{R}^{M_f\times d} $. The clip-scale branch downsamples the input into $M_c$ clips, followed by a FC layer and the \modelname{} block, producing clip embeddings $ \bm{V}_c = \{\bmc_i\}_{i=1}^{M_c} \in \mathbb{R}^{M_c \times d} $.  We unify the two embeddings under $\bmV_o \in \mathbb{R}^{M\times d}$, as they are processed identically.

After obtaining the optimal global registers $\bmr_0$, we leverage their global contextual semantics to enhance fine-grained video representations. Video features are  concatenated with the registers to form $\bmx = \mathrm{Concat}([\bmV_o, \bmr_0]) \in \mathbb{R}^{(M+N_r) \times d}$. A Register-Augmented Attention Block (RAB) then fuses them via a modified Gaussian attention \cite{GMMFORMER}, expressed as:
\begin{equation}
  \!\!\!\!\!  \text{GA}(\bmx) = \text{softmax}\left(\mathcal{M}_{r} + \left(\mathcal{M}_{\sigma}^{g} \odot \frac{ \bmx^q ( \bmx^k)^{\top}}{\sqrt{d_h}}\right) \right)\bmx^v,
     \label{eq:euclidean-attention}
\end{equation}
where $\mathcal{M}_{\sigma}^{g}$ is the Gaussian matrix applied only to video features. $\bmx^q, \bmx^k, \bmx^v$ are linear projections of $\bmx$, and $\odot$ denotes element-wise multiplication. As shown in \cref{fig:arc} (d), asymmetric attention masks $\mathcal{M}_r$ allow video tokens to attend to both registers and other video tokens, while registers interact only with video tokens.
The registers $\bmr_0$ are then discarded. Replacing the Transformer’s self-attention with Gaussian Attention constructs the Register-Augmented Attention Block. $N_a$ such blocks are arranged in parallel, whose outputs are aggregated via MAIM \cite{HLFormer} to form the \modelname{} block.

\subsection{Model Optimization}
Our framework is trained by optimizing the likelihood and enforcing register generation regularization to enhance representation learning, as defined in \cref{eq:optim}. For maximum likelihood estimation, we follow MS-SL \cite{ms-sl} and employ the standard similarity retrieval loss, denoted $L_{\mathrm{sim}}$, while the proposed losses specifically promote register generation. 
Finally, the total learning objective is:
\begin{equation}
    L_{\text{total}} = L_{\text{sim}}+L_{\text{tssl}}+L_{\text{pvs}}+\lambda_{dre}L_{\text{dre}}.
    \label{eq:total-loss}
\end{equation}

\begin{table*}[t]
\centering
\resizebox{\textwidth}{!}{
\begin{tabular}{l ccccc c ccccc c  ccccc}
\toprule
\multirow{2}{*}{\textbf{Model}}& \multicolumn{5}{c}{\textbf{ActivityNet Captions}} &  & \multicolumn{5}{c}{\textbf{Charades-STA}}&& \multicolumn{5}{c}{\textbf{TVR}} \\
\cmidrule{2-6} \cmidrule{8-12} \cmidrule{14-18}
& \textbf{R@1} & \textbf{R@5} & \textbf{R@10} & \textbf{R@100} & \textbf{SumR} && \textbf{R@1} & \textbf{R@5} & \textbf{R@10} & \textbf{R@100} & \textbf{SumR} & & \textbf{R@1} & \textbf{R@5} & \textbf{R@10} & \textbf{R@100} & \textbf{SumR} \\
\midrule
\midrule

\rowcolor{basegray}
\multicolumn{18}{l}{\scalebox{1.15}{\textit{Text-to-Video Retrieval (T2VR) Models}}}\\
RIVRL \cite{dong2022reading} & 5.2 & 18.0 & 28.2 & 66.4 & 117.8 &&1.6 & 5.6 & 9.4 & 37.7  & 54.3 && 9.4 & 23.4 & 32.2 & 70.6 & 135.6  \\
 DE++ \cite{dong2021dual}  & 5.3 & 18.4 & 29.2 & 68.0 & 121.0  && 1.7 & 5.6 & 9.6 & 37.1& 54.1 && 8.8 & 21.9 & 30.2 & 67.4 & 128.3\\
CLIP4Clip \cite{luo2022clip4clip} & 5.9 & 19.3 & 30.4 & 71.6 & 127.3 &&1.8 & 6.5 & 10.9 & 44.2 & 63.4 &&  9.9 & 24.3 & 34.3 & 72.5 & 141.0\\
Cap4Video  \cite{wu2023cap4video} & 6.3 & 20.4 & 30.9 & 72.6 & 130.2 &&1.9 & 6.7 & 11.3 & 45.0 & 65.0 &&  10.3 & 26.4 & 36.8 & 74.0 & 147.5\\
\midrule

\rowcolor{basegray}
\multicolumn{18}{l}{\scalebox{1.15}{ \textit{Video Corpus Moment Retrieval (VCMR) Models w/o Moment localization}}}\\
ReLoCLNet \cite{zhang2021video} & 5.7 & 18.9 & 30.0 & 72.0 & 126.6 &&1.2 & 5.4 & 10.0 & 45.6 & 62.3 & & 10.0 & 26.5 & 37.3 & 81.3  & 155.1 \\
XML  \cite{lei2020tvr} &  5.3 & 19.4 & 30.6 & 73.1 & 128.4 &&1.6 & 6.0 & 10.1 & 46.9 & 64.6 && 10.7 & 28.1 & 38.1 & 80.3  & 157.1 \\ 
CONQUER \cite{hou2021conquer}&  6.5 & 20.4 & 31.8 & 74.3  & 133.1 &&1.8 & 6.3 & 10.3 & 47.5  & 66.0 &&11.0 & 28.9 & 39.6 & 81.3  & 160.8 \\ 
JSG \cite{jsg}&  6.8 & 22.7 & 34.8 & 76.1  & 140.5 &&2.4 & 7.7 & 12.8 & 49.8  & 72.7 &&- & - & - & -  & -\\ 
\midrule

\rowcolor{basegray}
\multicolumn{18}{l}{\scalebox{1.15}{\textit{Partially Relevant Video Retrieval (PRVR) Models}}}\\
MS-SL \cite{ms-sl}& 7.1 & 22.5 & 34.7 & 75.8 & 140.1 &&1.8 & 7.1 & 11.8 & 47.7 & 68.4& & 13.5 & 32.1 & 43.4 & 83.4 & 172.4       \\
MS-SL++ \cite{MS-SL++}& 7.0 & 23.1 & 35.2 & 75.8 & 141.1 && 1.8 & 7.6 & 12.0 & 48.4 & 69.7 && 13.6 & 33.1 & 44.2 & 83.5 & 174.5\\
 PEAN \cite{PEAN}&7.4 & 23.0 & 35.5 & 75.9 & 141.8 &&  2.7 & 8.1 & 13.5 & 50.3 & 74.7 && 13.5 & 32.8 & 44.1 & 83.9 & 174.2 \\
  LH \cite{lh}& 7.4 & 23.5 & 35.8 & 75.8 & 142.4 &&2.1 & 7.5 & 12.9 & 50.1 & 72.7& & 13.2 & 33.2 & 44.4 & 85.5 & 176.3 \\
 BGM-Net \cite{bgmnet}& 7.2 & 23.8 & 36.0 & 76.9 & 143.9 &&1.9 & 7.4 & 12.2 & 50.1 & 71.6 && 14.1 & 34.7 & 45.9 & 85.2 & 179.9 \\
 GMMFormer \cite{GMMFORMER}&8.3 & 24.9 & 36.7 & 76.1 & 146.0 && 2.1 & 7.8 & 12.5 & 50.6 & 72.9 &&13.9 & 33.3 & 44.5 & 84.9 & 176.6 \\
 ProtoPRVR \cite{ProtoPRVR} & 7.9 & 24.9 & 37.2 & 77.3 & 147.4 && - & - & - & - & - && 15.4 & 35.9 & 47.5 & 86.4 & 185.1\\
 DL-DKD \cite{DL-DKD}& 8.0 & 25.0 & 37.5 & 77.1 & 147.6 &&- & - & - & - & - && 14.4 & 34.9 & 45.8 & 84.9 & 179.9  \\
ARL \cite{ARL}& 8.3 & 24.6 & 37.4 & 78.0 & 148.3 && - & - & - & - & - && 15.6 & 36.3 & 47.7 & 86.3 & 185.9\\
MGAKD \cite{MGAKD}& 7.9 & 25.7 & 38.3 & 77.8 & 149.6 && - & - & - & - & - && 16.0 & 37.8 & 49.2 & 87.5 & 190.5\\
GMMFormerV2 \cite{GMMFormerV2}& \textbf{8.9} & 27.1 & 40.2 & 78.7 & 154.9 && 2.5 & 8.6 & 13.9 & 53.2 & 78.2 && 16.2 & 37.6 & 48.8 & 86.4 & 189.1\\
HLFormer \cite{HLFormer}& 8.7 & 27.1 & 40.1 & 79.0 & 154.9 && 2.6 & 8.5 & 13.7 & 54.0 & 78.7 && 15.7 & 37.1 & 48.5 & 86.4 & 187.7\\
\rowcolor{ourscolor}
 \textbf{\modelname{} (ours)} & 8.7 & \textbf{27.5} & \textbf{40.3} & \textbf{79.5} & \textbf{156.1} &&2.6 & \textbf{8.7} & \textbf{14.5} & \textbf{54.2} & \textbf{80.0} &&   \textbf{17.4} & \textbf{39.0} & \textbf{50.4} & \textbf{86.2} & \textbf{193.1}     \\
\hline
\bottomrule
\end{tabular}
}
\vspace{-.8em}
\caption{\textbf{Retrieval performance comparison across three standard benchmarks.} Models are ranked in ascending order of SumR scores on ActivityNet Captions.
State-of-the-art performance is highlighted in \textbf{bold}, while “-” denotes unavailable results.}
\vspace{-1.5em}
\label{tab:main}
\end{table*}

\subsection{Cross-modal Similarity Computation}
To compute the similarity between a text-video pair $(Q, V)$, we first extract the embeddings $\bmq$, $\bm{V}_f$, and $\bm{V}_c$. Frame-level and clip-level scores are  obtained using cosine similarity with a max operation:
\begin{equation}
\begin{aligned}
S_f(Q, V) &= \max \{ \cos(\bmq, \bmf_1), \dots, \cos(\bmq, \bmf_{M_f}) \}, \\
S_c(Q, V) &= \max \{ \cos(\bmq, \bmc_1), \dots, \cos(\bmq, \bmc_{M_c}) \}.
\label{eq:simscore}
\end{aligned}
\end{equation}
The final text-video similarity is then computed as:
\begin{equation}
S(Q, V) = \alpha_f S_f(Q, V) + \alpha_c S_c(Q, V),
\label{eq:sim}
\end{equation}
where $\alpha_f, \alpha_c \in [0, 1]$  satisfy $\alpha_f + \alpha_c = 1$. Videos are  retrieved and ranked based on the final similarity score.

%% file: sections/Experiments.tex
\begin{table}[t]
    \centering
    \resizebox{\linewidth}{!}{
        \begin{tabular}{@{}lccccc@{}}
        \toprule
        \textbf{Model} & 
        \makecell{\textbf{Train} \\ \textbf{time/epoch (ms)}} & 
        \makecell{\textbf{Model} \\ \textbf{parameters (M)}} & 
        \makecell{\textbf{Inference} \\ \textbf{time (ms)}} & 
        \makecell{\textbf{Retrieval} \\ \textbf{time (ms)}} & 
        \textbf{SumR} \\
        \midrule
        GMMFormer & 26887 & 12.85 & 2876 & 3238 & 72.9 \\
        HLFormer & 31463 & 28.43  & 3816 & 3655 & 78.7 \\
        GMMFormerV2 & 38004 & 30.79  & 3843 & 3688 & 78.2 \\
        \modelname{} & 33609 & 36.14  & 4001 & 3686 & 80.0 \\
        \bottomrule
        \end{tabular}
    }
    \vspace{-.5em}
    \caption{\textbf{Training and Inference Efficiency.}  Inference time denotes feature extraction for 1334 videos in the evaluation sets, while retrieval time accounts for the total time of query encoding, similarity computation and ranking.  Dataset: Charades-STA.}
    \label{tab:efficiency}
\end{table}

\section{Experiments} \label{sec:experiments}
\subsection{Experimental Setup} \label{subsec:exp_setup}
\noindent \textbf{Datasets} \quad We evaluate \modelname{} on three benchmark datasets: \textbf{(i) ActivityNet Captions} \cite{krishna2017dense}, which features roughly 20K YouTube videos, characterized by an average duration of 118 seconds. On average, each video is annotated with 3.7 moments and paired with a textual description.  \textbf{(ii) Charades-STA} \cite{gao2017tall}, which includes 6,670 videos with 16,128 sentence descriptions, averaging  2.4 moments with textual queries. \textbf{(iii) TV show Retrieval (TVR)} \cite{lei2020tvr}, composed of 21.8K video clips sourced from six different TV shows. Five natural language descriptions are provided for each.  The same data split used in MS-SL \cite{ms-sl,GMMFORMER} is  adopted in our experiment. Moment annotations are unavailable.

\noindent \textbf{Metrics} \quad Following prior works \cite{ms-sl,GMMFORMER}, we employ rank-based metrics for evaluation, specifically $R$@$K$($K$ = 1, 5, 10, 100). $R$@$K$ is defined as the percentage of queries where the ground-truth item appears within the top $K$ ranked results. All results are presented in percentages ($\%$). We also report the Sum of Recalls (SumR)  for an overall evaluation.

\begin{table*}[t]
\centering
\resizebox{\textwidth}{!}{
\begin{tabular}{cl ccccc c ccccc c  ccccc}
\toprule
\multirow{2}{*}{ID}&\multirow{2}{*}{\textbf{Model}}& \multicolumn{5}{c}{\textbf{ActivityNet Captions}} &  & \multicolumn{5}{c}{\textbf{Charades-STA}}&& \multicolumn{5}{c}{\textbf{TVR}} \\
\cmidrule{3-7} \cmidrule{9-13} \cmidrule{15-19}
&& \textbf{R@1} & \textbf{R@5} & \textbf{R@10} & \textbf{R@100} & \textbf{SumR} && \textbf{R@1} & \textbf{R@5} & \textbf{R@10} & \textbf{R@100} & \textbf{SumR} & & \textbf{R@1} & \textbf{R@5} & \textbf{R@10} & \textbf{R@100} & \textbf{SumR} \\
\midrule
\midrule
\rowcolor{ourscolor}
(0)& \textbf{\modelname{} (full)} & \textbf{8.7} & \textbf{27.5} & \textbf{40.3} & \textbf{79.5} & \textbf{156.1} &&\textbf{2.6} & \textbf{8.7} & \textbf{14.5} & \textbf{54.2} & \textbf{80.0} &&   \textbf{17.4} & \textbf{39.0} & \textbf{50.4} & \textbf{86.2} & \textbf{193.1}      \\
\midrule
\rowcolor{basegray}
\multicolumn{19}{l}{\scalebox{1.1}{\textit{Efficacy of Register Generation Strategy}}}\\
(1)& $w/o$ registers & 8.5& 26.6 & 39.5 & 78.8 & 153.4&&2.4 & 7.9 & 13.6 & 52.9  & 76.8 &&  15.7& 36.6 & 48.7&86.0 & 187.0 \\
(2)& $w/$ AP & 8.5& 26.2 & 39.3 & 77.9 & 151.9&&2.6 & 8.6 & 14.4 & 52.6 & 78.1 &&  17.0& 38.3 & 49.6&86.4 & 191.4 \\
(3)& $w/o$ DRE & 8.2& 25.7 & 38.7 & 78.0 & 150.6&&2.4 & 8.5 & 14.8 & 52.6  & 78.3 &&  16.8& 38.2 & 49.6&86.3 & 190.8 \\
(4)& $w/o$ PVS & 8.4& 27.3 & 40.2 & 78.9 & 154.9&&2.0 & 8.3 & 13.9 & 53.4  & 77.6 &&  16.5& 38.5 & 49.4&86.4 & 190.9 \\
\midrule
\rowcolor{basegray}
\multicolumn{19}{l}{\scalebox{1.1}{\textit{Efficacy of Different Loss Terms}}}\\
(5) & $L_{\text{sim}}$ Only & 8.1 & 25.4 & 38.3 & 78.7 & 150.5 &&2.4 & 7.8 & 13.3 & 53.1 & 76.6 & & 15.9 & 36.7& 48.3 & 86.1  & 187.0\\
(6) &$w/o$ $ L_{\text{pvs}}$&  8.6 & 27.3 & 40.1 & 79.1 & 155.1  &&2.3 & 8.5 & 14.0 & 53.7  & 78.5 &&16.9 & 38.4 & 49.7 & 86.7  & 191.6\\ 
(7) &$w/o$ $ L_{\text{dre}}$&  8.5 & 26.6 & 39.2 & 78.9 & 153.2  &&2.0 & 8.7 & 14.3 & 53.4  & 78.5 &&16.5 & 38.2 & 49.7 & 86.5  & 190.9\\ 
(8) & $w/o$ $L_{\text{tssl}}$  &  8.2 & 25.7 & 38.6 & 78.8 & 151.3 &&1.7 & 8.1 & 13.6 & 53.4 & 76.9 && 16.9 & 38.1 & 49.6 & 86.6  & 191.1\\ 
\hline
\bottomrule
\end{tabular}
}
\vspace{-.5em}
 \caption{Ablation Study of \modelname{}.  The best scores are marked in \textbf{bold}.}
 \label{tab:ablation}
\end{table*}
\begin{figure}[t]
\centering
    \vspace{-.5em}
    {
        \includegraphics[width = 0.2200\textwidth]{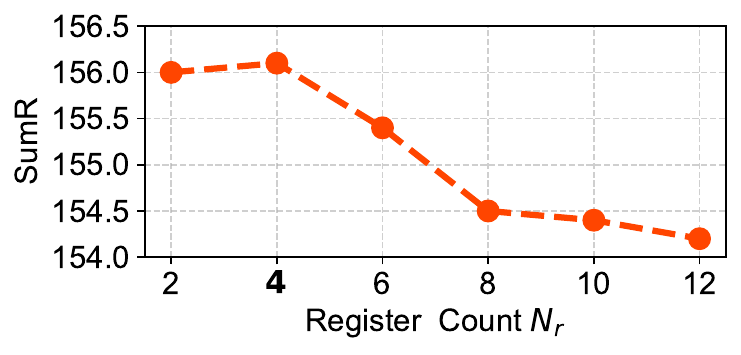}
    }
    {
        \includegraphics[width = 0.2200\textwidth]{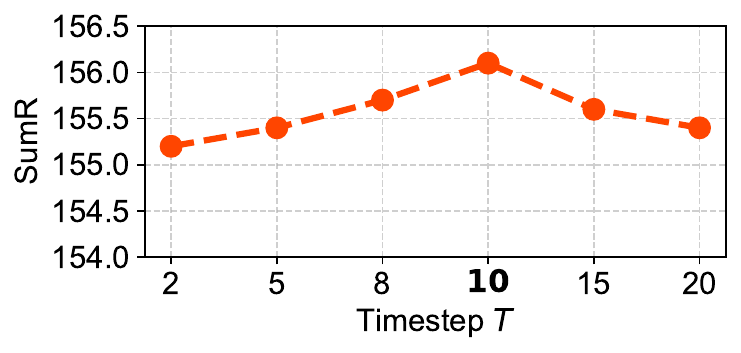}
    }\\
        \vspace{-0.5em}
    \scalebox{0.8}{\small {(a) ActivityNet Captions}} \\
    {
        \includegraphics[width = 0.2200\textwidth]{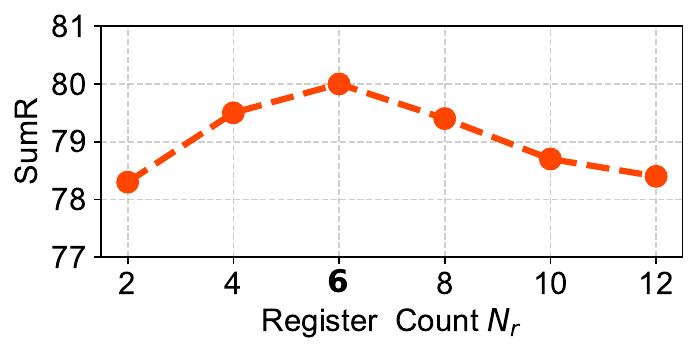}
    }
    {
        \includegraphics[width = 0.2200\textwidth]{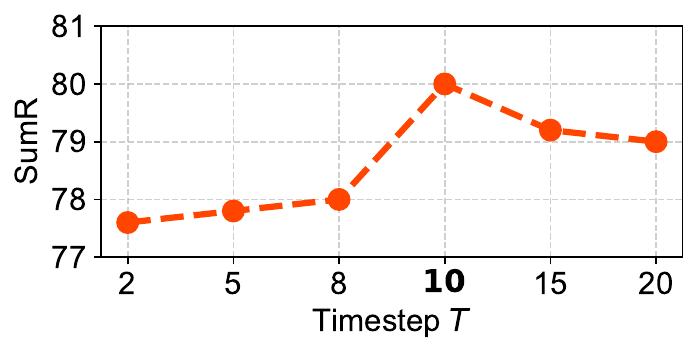}
    }\\
        \vspace{-0.5em}
    \scalebox{0.8}{{(b) Charades-STA}} \\
    {
        \includegraphics[width = 0.2200\textwidth]{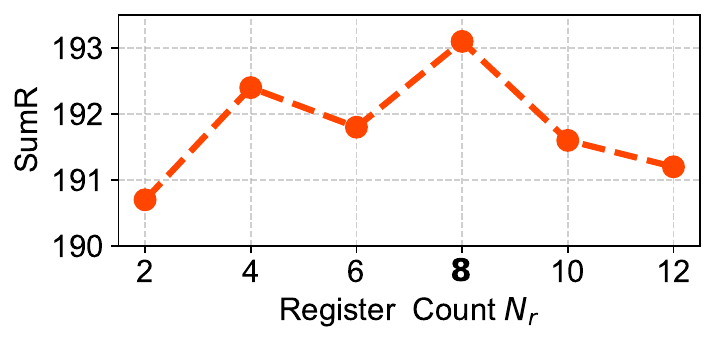}
    }
    {
        \includegraphics[width = 0.2200\textwidth]{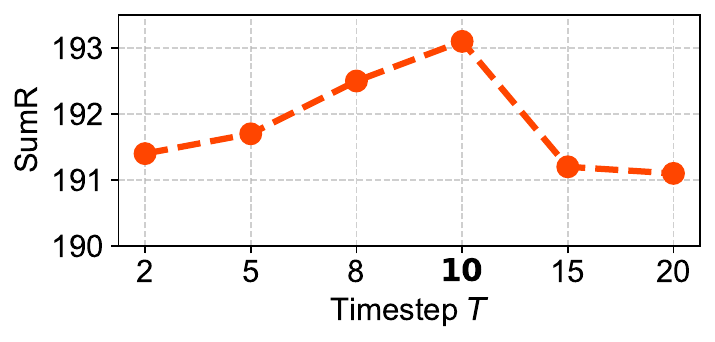}
    }\\
    \vspace{-0.5em}
    \scalebox{0.9}{\small {(c) TVR}} \\
    \vspace{-1em}
    \caption{The influence of the number of registers  and the number of diffusion timesteps, with default settings marked in bold.}
    \vspace{-1em}
    \label{fig:ablation-reg-num}
\end{figure}

\subsection{Implementation Details}
\noindent \textbf{Data Pre-Processing} \quad For both the ActivityNet Captions and Charades-STA, we use the provided I3D features for video representations obtained by \citet{zhang2020hierarchical} and \citet{mun2020local}, respectively. Additionally, we employ 1,024-dimensional RoBERTa features extracted via MS-SL \cite{ms-sl} for query representations. For TVR, we utilize the 3,072-dimensional video features provided by \citet{lei2020tvr}, which concatenate frame-level ResNet152 \cite{he2016deep} and segment-level I3D representations \cite{carreira2017quo}. The corresponding textual data is processed into the 768-dimensional RoBERTa \cite{liu2019roberta} features. 

\noindent \textbf{Experimental Configurations} \quad 
The \modelname{} block consists of 8 Register-augmented 
Attention Blocks  ($N_a$ = 8), with Gaussian variances ranging from $2^{-2}$ to $2^{N_{a}-3}$ and $\infty$. The latent dimension  $d$ = 384 with 4 attention heads.
For the diffusion timesteps, we set $T$ = 10.  $N_r$ is set to 6 for Charades-STA, 4 for ActivityNet Captions, and 8 for TVR. 
The model is implemented in PyTorch and trained on a single Nvidia A100-40G GPU. We employ Adam \cite{kingma2014adam} as the optimizer and set the mini-batch size to 128.

\subsection{Comparison with State-of-the arts}
\noindent \textbf{Baselines} \quad We select twelve representative PRVR baselines for comparison.
In addition, we evaluate   \modelname{} against  other methods in T2VR and VCMR. For T2VR,
 four models are included:
  RIVRL \cite{dong2022reading}, DE++ \cite{dong2021dual}, CLIP4Clip \cite{luo2022clip4clip} and Cap4Video \cite{wu2023cap4video}.
For VCMR, we compare with  ReLoCLNet \cite{zhang2021video}, XML \cite{lei2020tvr},  CONQUER \cite{hou2021conquer} and JSG \cite{jsg}.

\noindent \textbf{Retrieval Performance} \quad
We present retrieval performance of various models in \cref{tab:main}. PRVR models tailored specifically for this task achieve the best performance, surpassing both T2VR and VCMR models. Distinct among PRVR models, \modelname{} leverages the  global registers to achieve better retrieval capabilities, demonstrating a clear advantage over all existing baseline methods. 
Its inherent generation capability alleviates the problem of incomplete global contextual perception, enabling it to extract reliable holistic semantics, ultimately improving retrieval performance.

\noindent \textbf{Model Efficiency} \quad
We further evaluate the retrieval efficiency  by measuring model parameters, the training time per epoch, the video feature extraction time and the overall retrieval time. All results are averaged over 10 runs under the same experimental environment. As shown in \cref{tab:efficiency}, our model achieves comparable efficiency to HLFormer, with a slight overhead introduced by the iterative diffusion-based register generation. Nevertheless, this trade-off yields performance gains and both training and inference times are acceptable, demonstrating the high efficiency of our framework. In offline scenarios, since video features are precomputed and cached, generating global registers in the video branch imposes negligible additional cost during retrieval.
\begin{figure}[t]
\centering
\vspace{-.8em}
    \setcounter{subfigure}{0}
  \subfloat[\small{$L_{\text{div}}$ Only}]{\includegraphics[width = 0.2300\textwidth]{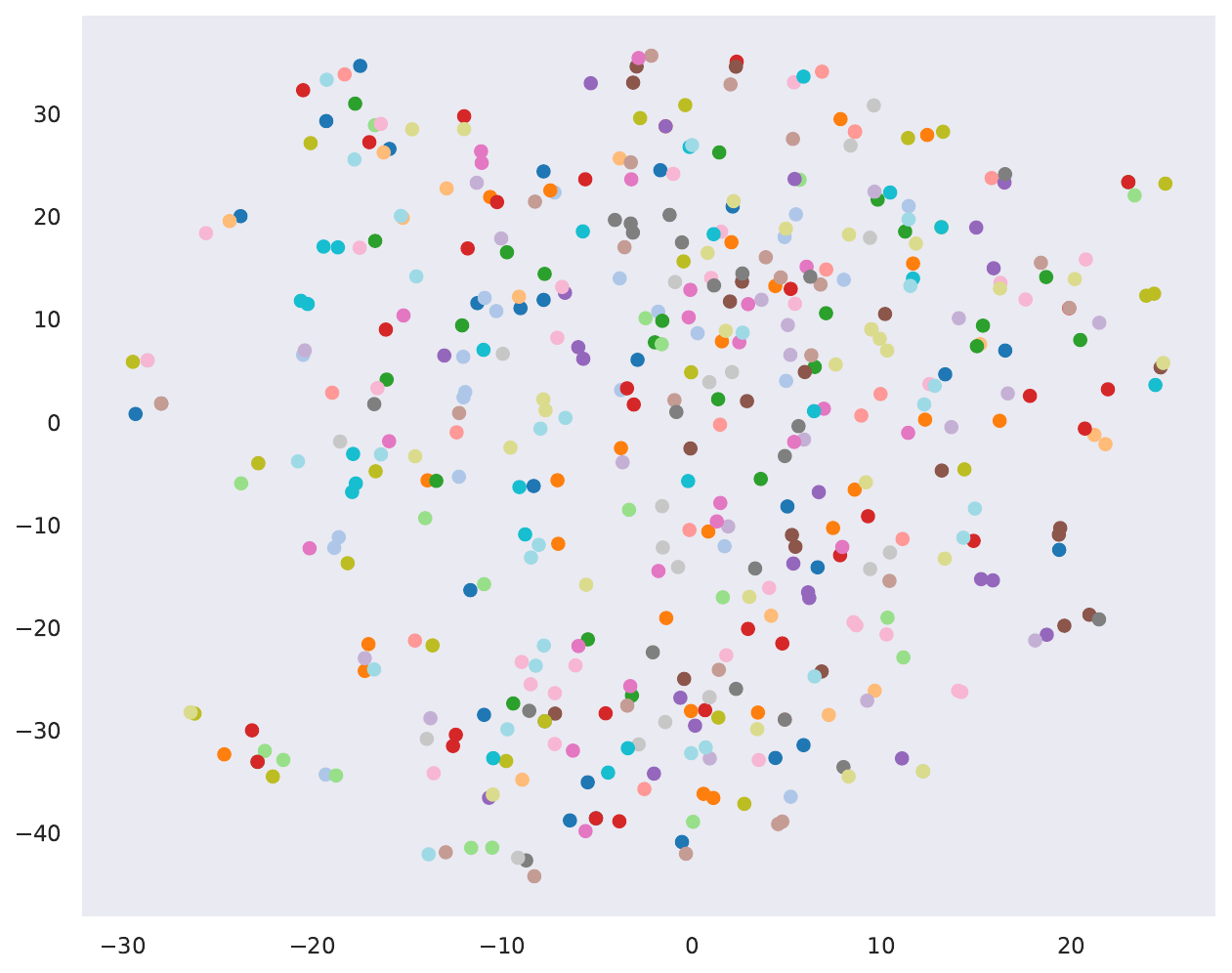}}
  \subfloat[\small{$L_{\text{tssl}}$ Full}]{\includegraphics[width = 0.2300\textwidth]{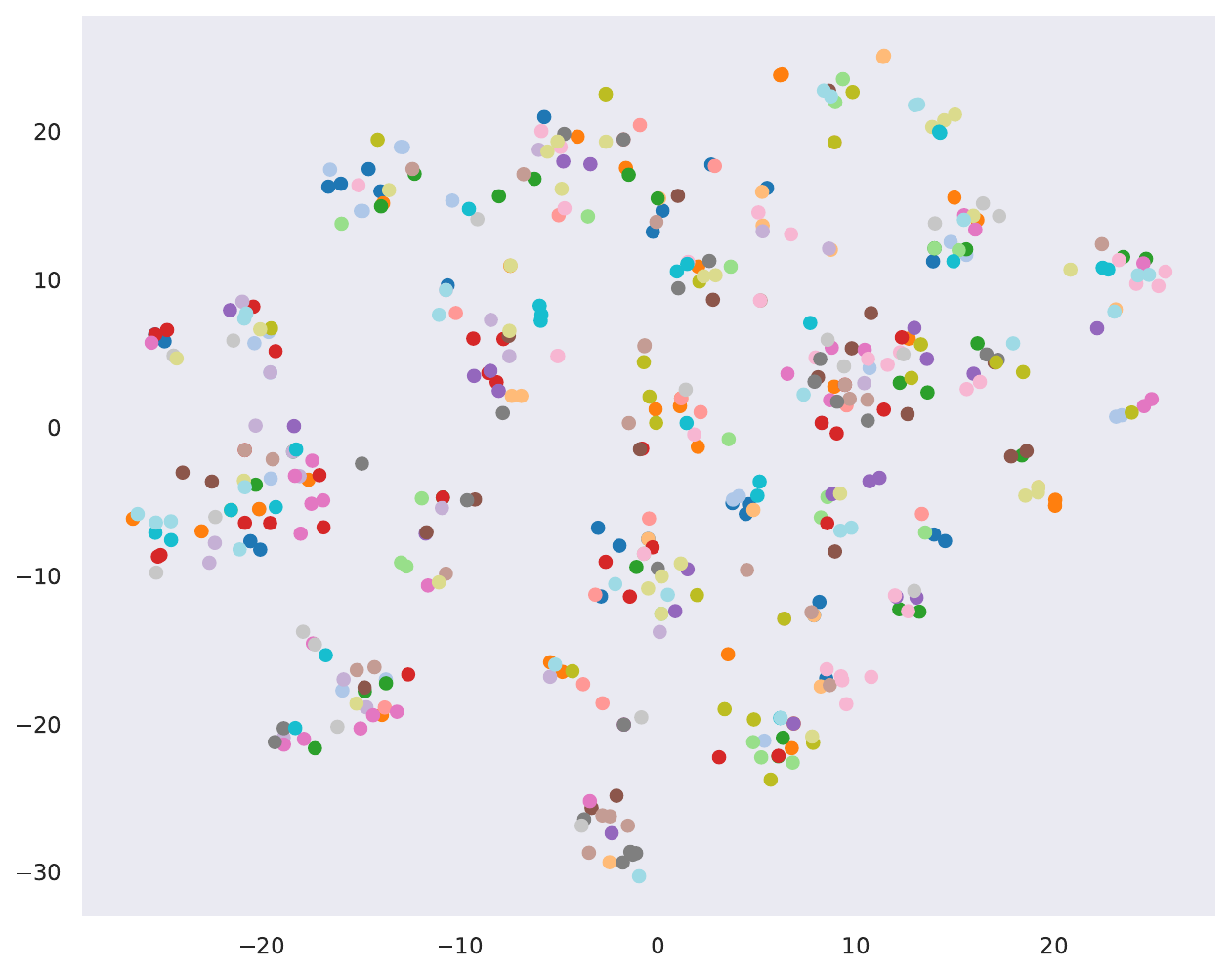}}
  \vspace{-.8em}
  \caption{The t-SNE visualization of the learned textual space. Data points of the  same color denote queries from the same video. }
  \label{fig:query-tsne}
\end{figure}
\subsection{Model Analyses} 
\noindent \textbf{Effects of the number of registers $N_r$ and diffusion timesteps $T$} \quad
We conduct ablation studies on register quantity and diffusion timesteps, as shown in \cref{fig:ablation-reg-num}.
Very few registers yield suboptimal performance due to insufficient capacity to capture holistic video semantics, whereas excessive registers ($N_r>8$) may introduce redundancy and degrade performance.
Empirically, employing 4--8 registers yields robust results across all three datasets.
As for the diffusion timesteps $T$, performance steadily improves as $T$ increases from 2 to 10, peaking at $T=10$, which highlights the importance of iterative refinement. 
 Then, performance uniformly declines when $T > 10$, suggesting over-refinement and potential overfitting to textual supervision.
Balancing accuracy and efficiency, we set $T=10$ as default. The effectiveness achieved with relatively few registers and timesteps further underscores the high efficiency of our framework.

\noindent \textbf{Efficacy of Register Generation Strategy} \quad 
\cref{tab:ablation} compares four register generation strategies: (\textbf{i}) \textbf{w/o registers}: omitting registers degrades performance, highlighting the value of global contextual information; (\textbf{ii}) \textbf{w/ $\mathbf{AP}$}: replacing the generative mechanism with adaptive pooling yields inferior results, indicating that simple pooling from untrimmed features is insufficient to capture meaningful global semantics; (\textbf{iii}) \textbf{w/o DRE}: one-step mapping without diffusion-based refinement performs worse, emphasizing the necessity of progressive refinement; (\textbf{iv}) \textbf{w/o PVS}: initializing registers from random Gaussian noise remains suboptimal, confirming the benefit of video-centric initialization.
\begin{figure}[t]
    \centering
    \begin{subfigure}[t]{0.48\textwidth}
        \includegraphics[width=\textwidth]{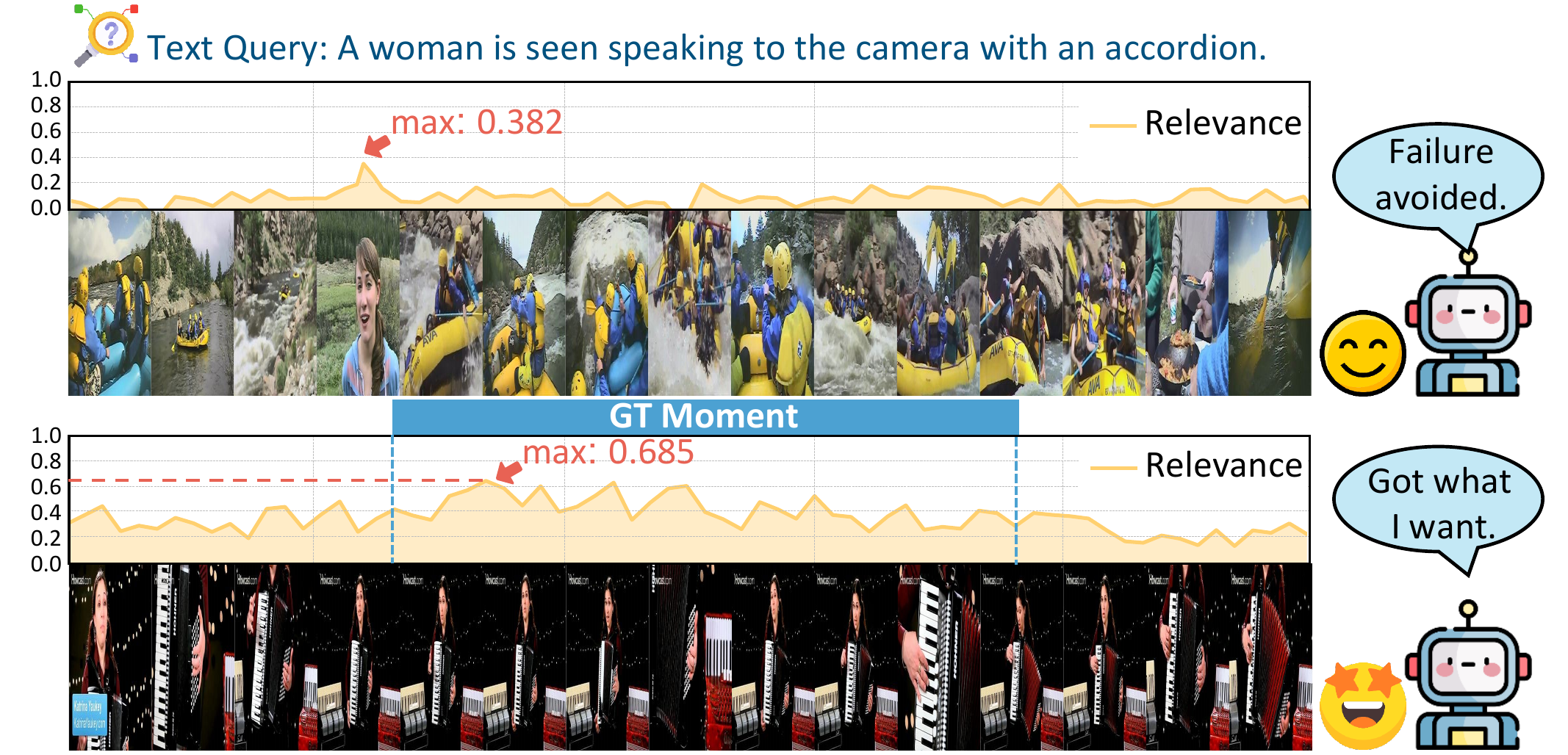}
    \end{subfigure}
    \vspace{-2em}
    \caption{A qualitative case study of retrieval results. The same videos in \cref{fig:Intro} are selected for a better comparison.}
    \label{fig:vis1}
    \vspace{-1em}
\end{figure}

\noindent \textbf{Effects of Different Loss Terms} \quad 
To analyze the contributions of the four losses in \cref{eq:total-loss}, we consider several variants:  
(\textbf{i}) \textbf{$L_{\text{sim}}$ Only}:  trained solely with $L_{\text{sim}}$.
(\textbf{ii}) \textbf{w/o $L_{\text{pvs}}$}: no constraint is imposed on the PVS sampling space.
(\textbf{iii}) \textbf{w/o $L_{\text{dre}}$}: registers are generated without textual supervision, resulting in unguided generation. 
(\textbf{iv}) \textbf{w/o $L_{\text{tssl}}$}: textual semantic structure learning is omitted.
As shown in \cref{tab:ablation}, the worst performance occurs when only $L_{\text{sim}}$ is used. Comparing Variant (0) with Variant (6), adding  $L_{\text{pvs}}$  increases the SumR, which can verify its necessity. 
Comparing Variant (0) with Variant (7), adding  $L_{\text{dre}}$ leverages textual supervision to ensure registers capture the holistic semantics conveyed by the text, thereby enhancing retrieval performance.  
Comparing Variant (0) with Variant (8)  and \cref{fig:query-tsne},  integrating $L_{\text{tssl}}$ not only boosts retrieval accuracy but also shapes a well-structured textual latent space.

\noindent \textbf{Visualization of Textual Latent Space} \quad
To gain deeper insights, we apply t-SNE \cite{tsne} to visualize the learned textual latent space, where a subset of videos and their corresponding queries from Charades-STA are randomly sampled for illustration. As depicted in \cref{fig:query-tsne}, using $L_{\text{div}}$ alone promotes semantic dispersion, enriching the latent semantics yet resulting in a scattered representation space. By introducing $L_{\text{qsp}}$, the full model maintains query semantic coherence and forms a more distinctive and well-structured latent manifold, which further provides effective semantic guidance for register generation and consequently facilitates retrieval.

\noindent \textbf{Efficacy of Global Registers} \quad In \cref{fig:vis1}, we visualize the temporal attention scores between queries and videos to examine the role of global registers. Compared with \cref{fig:Intro} (a), our model exhibits lower responses to incorrect videos while suppressing local spikes. Meanwhile, it produces higher similarity scores on the relevant video, with peak regions accurately aligned with the ground-truth moment. This confirms that registers provide global contextual guidance, enabling the model to suppress irrelevant content, mitigate spurious local noise and strengthen responses to relevant videos.

\begin{figure}[t]
\centering
    \setcounter{subfigure}{0}
  \subfloat[\small{$T$=0}]{\includegraphics[width = 0.1600\textwidth]{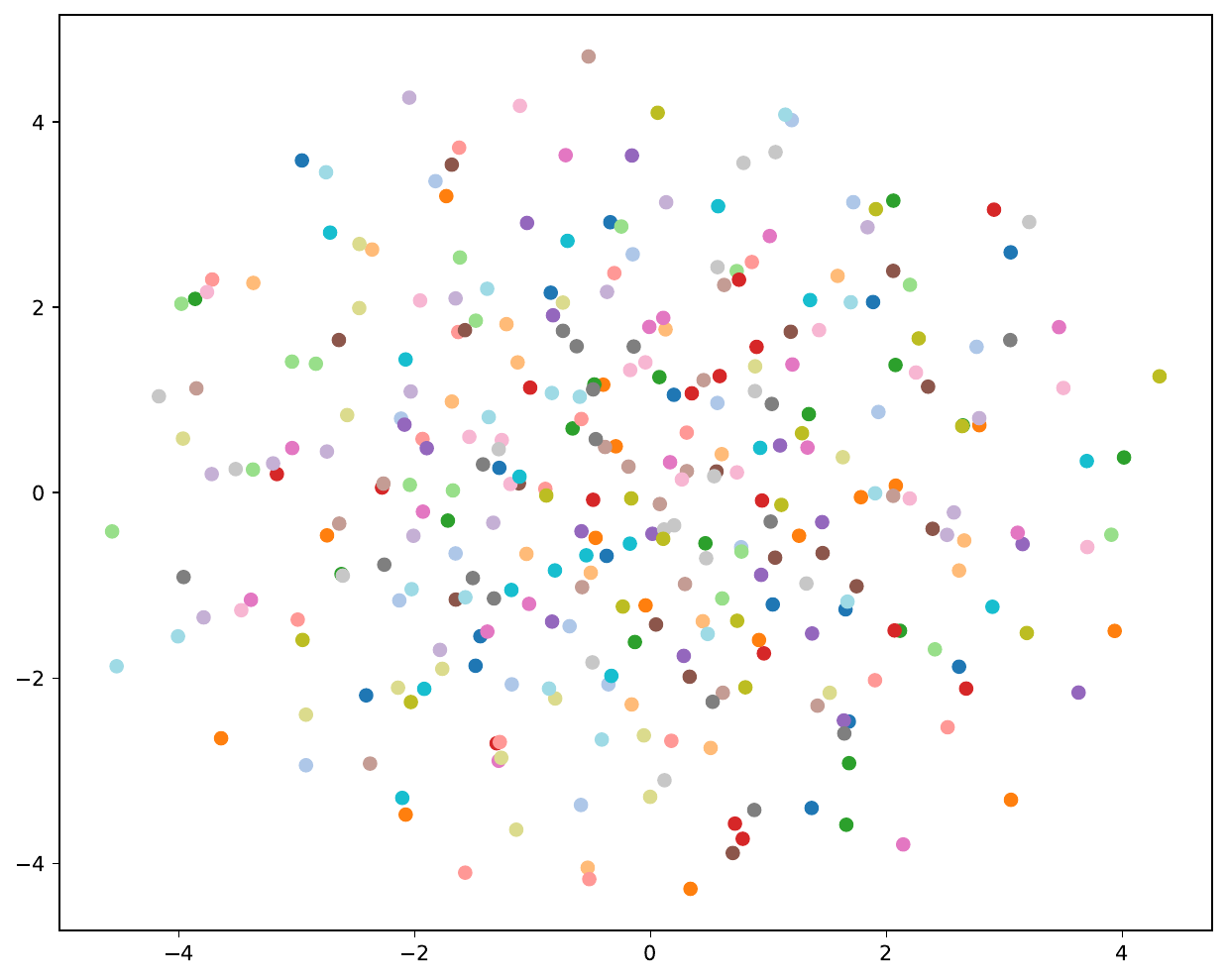}}
  \subfloat[\small{$T$=5}]{\includegraphics[width = 0.1600\textwidth]{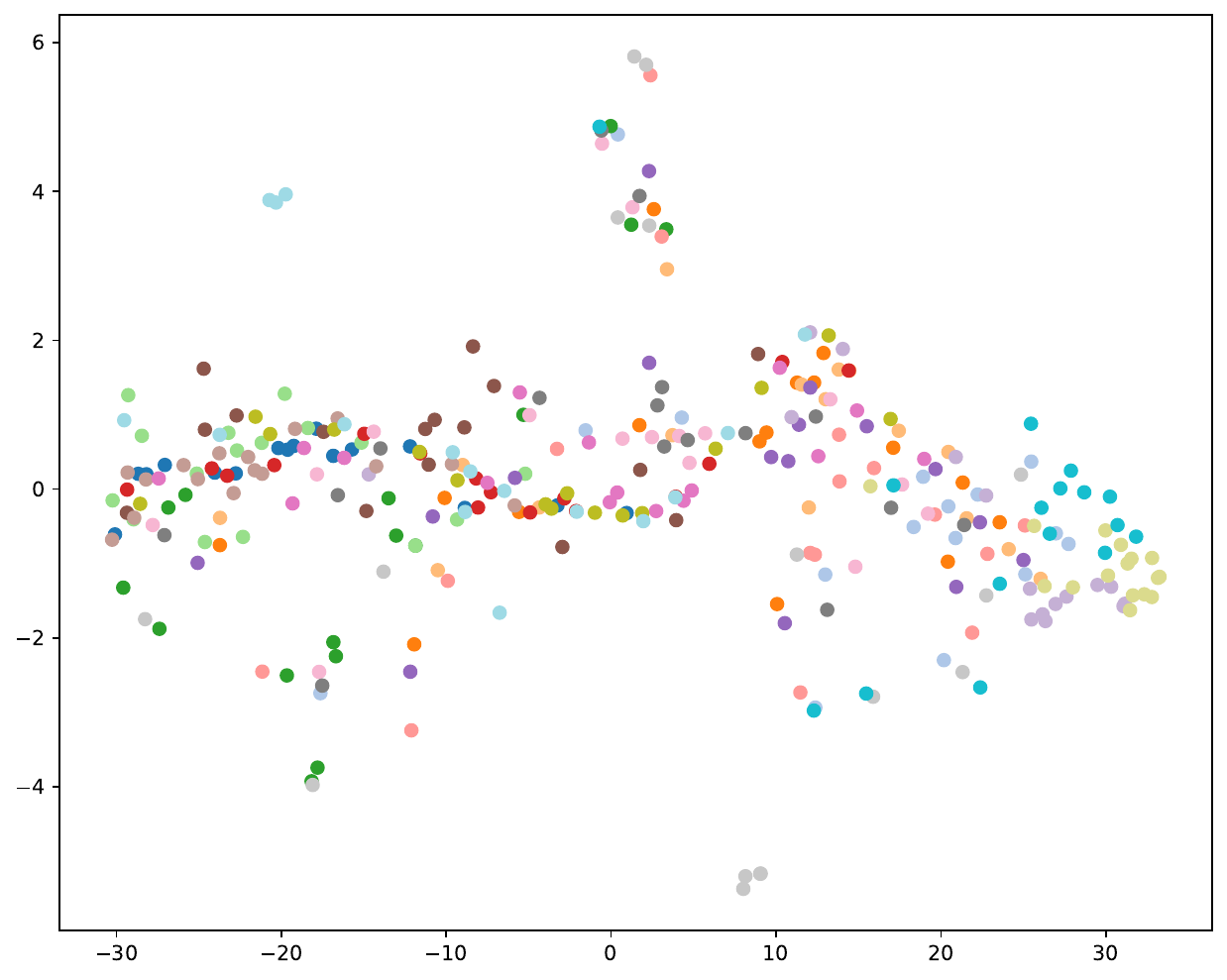}}
    \subfloat[\small{$T$=10}]{\includegraphics[width = 0.1600\textwidth]{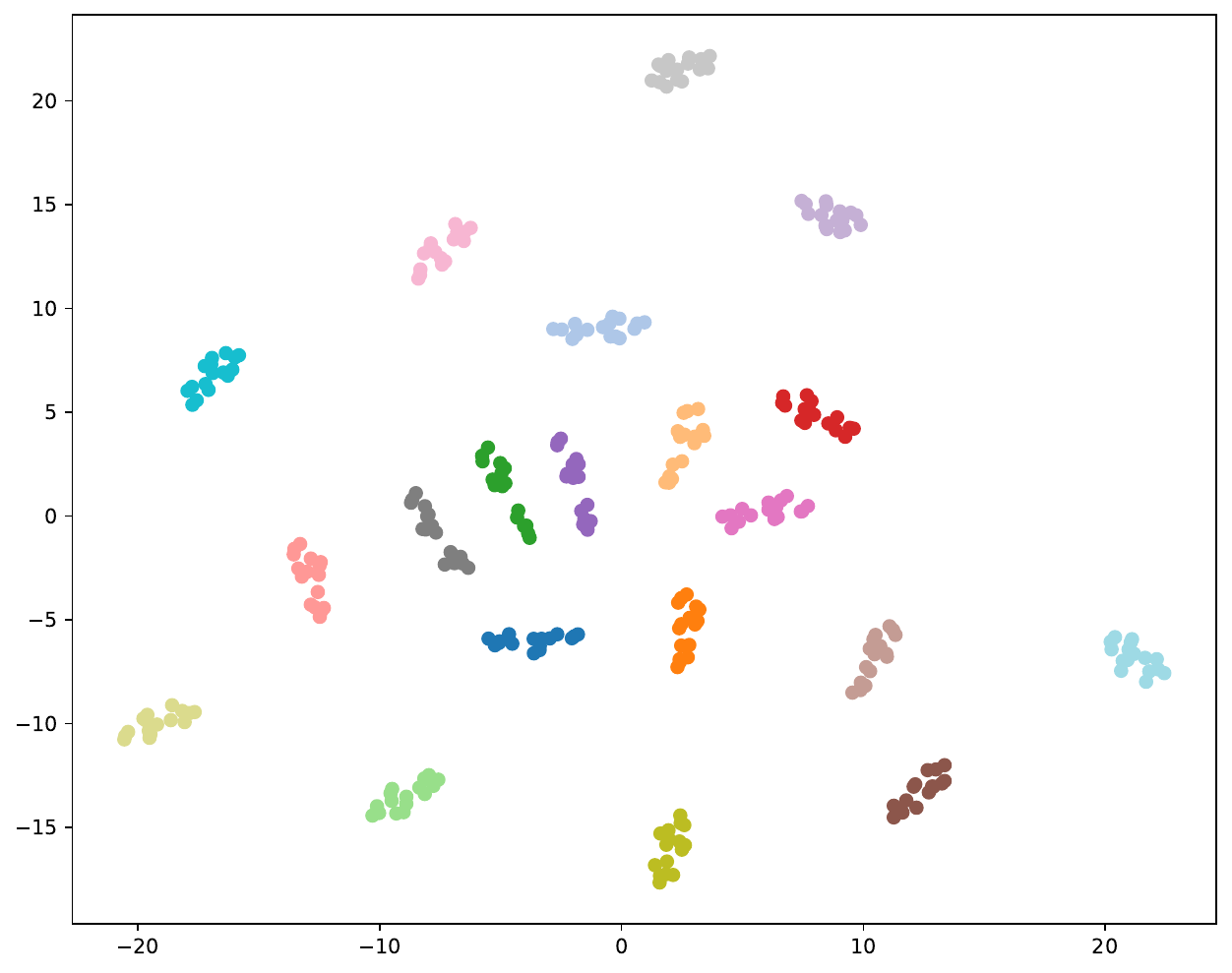}}
  \vspace{-.8em}
  \caption{The t-SNE visualization of the iterative generation process of global registers. $T$ = 0 indicates the initialization while $T$ = 10 represents the final results. 16 registers are trained for better clarity and data points of the same color correspond to  the same video.}
  \label{fig:register-tsne}
\end{figure}
\noindent  \textbf{Visualization of Register Generation} \quad To further illustrate the iterative generation process of registers, we randomly select 20 videos from Charades-STA and apply t-SNE to visualize the register space, as shown in \cref{fig:register-tsne}. The registers exhibit no clear semantics at initialization, but their representations become progressively purified through iterative refinement and denoising, eventually forming more discriminative clusters with clear video boundaries. This indicates that the registers capture reliable global video semantics and highlights the necessity of the diffusion process.

%% file: sections/Conclusions.tex
\section{Conclusions}
\label{sec:conclusion}
In this paper, we propose \modelname{}, an efficient diffusion-guided framework  for PRVR. Our model first generates coarse-grained global registers through a truncated diffusion process initialized from the video-centric distribution, capturing global semantics and enhancing fine-grained representation learning for better retrieval, realizing progressive hierarchical cross-modal alignment. Concurrently, textual semantic structure learning constructs a well-formed space, which provides stable and strong supervision for reliable generation.
Extensive experiments indicate that \modelname{}
outperforms state-of-the-art methods. Moreover, our approach introduces a unified generative–discriminative paradigm for PRVR, offering a new perspective which we hope can inspire future research.

\paragraph{Acknowledgments}
We sincerely thank the anonymous reviewers and chairs for their efforts and constructive suggestions, which have greatly helped us improve the manuscript. 
This work is supported in part by the National Natural Science Foundation of China under grants 624B2088, 62571298, 62576122, 62301189, and in part by the project of Peng Cheng Laboratory (PCL2025A14).

%% file: appendix_sections/Method.tex
\section{More Details on Method}
\subsection{Derivation of ELBO in Eq.~(2)}
Given the following predictive function:
\vspace{-.6em}
\begin{equation}
    p_{\theta,\phi}(Q|V) = \int \underbrace{p_\theta(Q|V,r)}_{\text{concentration}} \cdot 
    \underbrace{p_\phi(r|V)}_{\text{imagination}}  dr,
    \label{eq:appendix-register}
    \vspace{-.6em}
\end{equation}
where $p_{\phi}(r | V)$ denotes the video branch that first generates global registers, and $p_{\theta}(Q|V, r)$ models the register-augmented cross-modal matching. Our training objective is to maximize $p_{\theta,\phi}(Q|V)$. We introduce a variational posterior $q_{\varphi}(r | Q_a)$ to approximate the  true posterior $p(r | Q_a)$.

First, we rewrite \cref{eq:appendix-register} by introducing $q_{\varphi}(r|Q_a)$:
\vspace{-.6em}
\begin{equation}
    \begin{aligned}
    &\log p_{\theta,\phi}(Q|V)\\
    =& \log \int p_\theta(Q|V, r) p_\phi(r|V) dr  \\
=& \log \int p_\theta(Q|V, r) \frac{p_\phi(r|V)}{q_{\varphi}(r | Q_a)}q_{\varphi}(r | Q_a) dr.
    \end{aligned}
    \vspace{-.6em}
\end{equation}
Next, we invoke Jensen's inequality \cite{jensen1906fonctions}, leveraging the concavity of the $\log$ function, which satisfies:
\vspace{-.6em}
\begin{equation}
  \log \mathbb{E}[X] \geq \mathbb{E}[\log X].  
  \vspace{-.6em}
\end{equation}
Thus, the logarithm can be moved outside the integral, yielding a tractable lower bound:
\vspace{-.6em}
\begin{equation}
    \begin{aligned}
     & \log p_{\theta,\phi}(Q|V)\\
     \geq& \int q_{\varphi}(r | Q_a) \log \left( p_\theta(Q|V, r) \frac{p_\phi(r|V)}{q_{\varphi}(r | Q_a)} \right) dr. 
    \end{aligned}
    \vspace{-.6em}
\end{equation}
We subsequently decompose the logarithmic term within the integrand:
\vspace{-.6em}
\begin{equation}
    \begin{aligned}
    &\log \left( p_\theta(Q|V, r) \frac{p_\phi(r|V)}{q_{\varphi}(r | Q_a)} \right) \\
    =& \log p_\theta(Q|V, r) + \log \frac{p_\phi(r|V)}{q_{\varphi}(r | Q_a)}. 
    \end{aligned}
    \vspace{-.6em}
\end{equation}
Hence, the lower bound becomes:
\vspace{-.6em}
\begin{equation}
    \begin{aligned}
\log p_{\theta,\phi}(Q|V) \geq & \int q_{\varphi}(r | Q_a) \log p_\theta(Q|V, r) dr \\
& - \int q_{\varphi}(r | Q_a) \log \frac{q_{\varphi}(r | Q_a)}{p_\phi(r|V)} dr.   
    \end{aligned}
    \vspace{-.6em}
\end{equation}
Therefore, we obtain the final form of the ELBO in Eq.~(2):
\vspace{-.6em}
\begin{equation}
    \begin{aligned}
        \log p_{\theta, \phi}(Q | V) &\geq \mathbb{E}_{q_\varphi(r | Q_a)} \left[ \log p_\theta(Q | V, r) \right] \\
        &\quad - \mathbb{KL} \left[ q_\varphi(r | Q_a) \, \| \, p_\phi(r | V) \right].
    \end{aligned}  
    \label{eq:appendix-elbo}
        \vspace{-.6em}
\end{equation}

\subsection{Relationship between Eq.~(7) and $L_{\text{dre}}$}
Following ~\cite{luo2022understanding}, Bayes’ rule gives:
\vspace{-.6em}
\begin{equation}
    q(\bm{\hat{q}}_t | \bm{\hat{q}}_{t-1}, \bm{\hat{q}}_0) = \frac{q(\bm{\hat{q}}_{t-1} | \bm{\hat{q}}_t, \bm{\hat{q}}_0) q(\bm{\hat{q}}_t | \bm{\hat{q}}_0)}{q(\bm{\hat{q}}_{t-1} | \bm{\hat{q}}_0)}.
    \label{eq:tool}
    \vspace{-.6em}
\end{equation}
Armed with this new equation, we can retry the derivation resuming from the ELBO in Eq.~(7) by viewing $\bm{\hat{q}}$  as $\bm{\hat{q}}_0$:
\vspace{-.6em}
\begin{equation}\label{eq:DM-ELBO}
\small
\begin{aligned}
&\log p({\bm{\hat{q}}}_0)\\
=&\log\int p({\bm{\hat{q}}}_{0:T})\mathrm{d}{\bm{\hat{q}}}_{1:T} \\
=&\log\mathbb{E}_{q({\bm{\hat{q}}}_{1:T}|{\bm{\hat{q}}}_0)}\left[\dfrac{p({\bm{\hat{q}}}_{0:T})}{q({\bm{\hat{q}}}_{1:T}|{\bm{\hat{q}}}_0)}\right] \\
\ge& \underbrace{\mathbb{E}_{q({\bm{\hat{q}}}_1|{\bm{\hat{q}}}_0)}\left[\log p_\phi({\bm{\hat{q}}}_0|{\bm{\hat{q}}}_1)\right]}_{\small(\text{reconstruction term})} - \underbrace{\mathbb{KL}(q({\bm{\hat{q}}}_T|{\bm{\hat{q}}}_0)\parallel p({\bm{\hat{q}}}_T))}_{\small(\text{prior matching term})} \\
-& \textstyle\sum_{t=2}^{T}\underbrace{\mathbb{E}_{q({\bm{\hat{q}}}_t|{\bm{\hat{q}}}_0)}\left[\mathbb{KL}(q({\bm{\hat{q}}}_{t-1}|{\bm{\hat{q}}}_t,{\bm{\hat{q}}}_0)\parallel p_\phi({\bm{\hat{q}}}_{t-1}|{\bm{\hat{q}}}_t))\right]}_{\small(\text{denoising matching term})},
\end{aligned} 
\vspace{-.6em}
\end{equation}
where (i) the reconstruction term corresponds to the negative reconstruction error over ${\bm{\hat{q}}}_0$; (ii) the prior matching term is constant with no trainable parameters and can thus be ignored during optimization; and (iii) the denoising matching terms constrain $p_\phi({\bm{\hat{q}}}_{t-1} \mid {\bm{\hat{q}}}_t)$ to align with the tractable ground-truth transition $q({\bm{\hat{q}}}_{t-1} \mid {\bm{\hat{q}}}_t, {\bm{\hat{q}}}_0)$~\cite{luo2022understanding}. Consequently, $\phi$ is optimized to iteratively recover ${\bm{\hat{q}}}_{t-1}$ from ${\bm{\hat{q}}}_t$. 
Following~\cite{DDPM}, the denoising matching terms can be simplified as
\vspace{-.7em}
\begin{equation}
\sum_{t=2}^{T} \mathbb{E}_{t,\bm{\epsilon}} \Big[ \| \bm{\epsilon} - \bm{\epsilon}_\phi({\bm{\hat{q}}}_t,t) \|_2^2 \Big],
\vspace{-.6em}
\label{eq:appendix-standard-elbo}
\end{equation}
where $\bm{\epsilon} \sim \mathcal{N}(\bm{0},\bm{I})$, and $\bm{\epsilon}_\phi({\bm{\hat{q}}}_t,t)$ is parameterized by a neural network (e.g., U-Net~\cite{DDPM}) to predict the noise $\bm{\epsilon}$ that generates ${\bm{\hat{q}}}_t$ from ${\bm{\hat{q}}}_0$ in the forward process~\cite{luo2022understanding}. A detailed derivation of Eq.~(7) and \cref{eq:DM-ELBO} is provided in \cite{luo2022understanding}.

$L_{\text{dre}}$ extends \cref{eq:appendix-standard-elbo} by incorporating a conditioning variable $\bmc$. Inspired by ~\cite{DDPM}, the reconstruction term in \cref{eq:DM-ELBO} has a relatively minor effect; hence, $L_{\text{dre}}$ is employed to optimize the denoising matching terms in \cref{eq:DM-ELBO}, thereby providing an approximate estimation of \cref{eq:DM-ELBO} for training.

\subsection{Further details of \modelname{} architecture}
\begin{figure}[t]
    \centering
    \begin{subfigure}[t]{0.48\textwidth}
        \includegraphics[width=\textwidth]{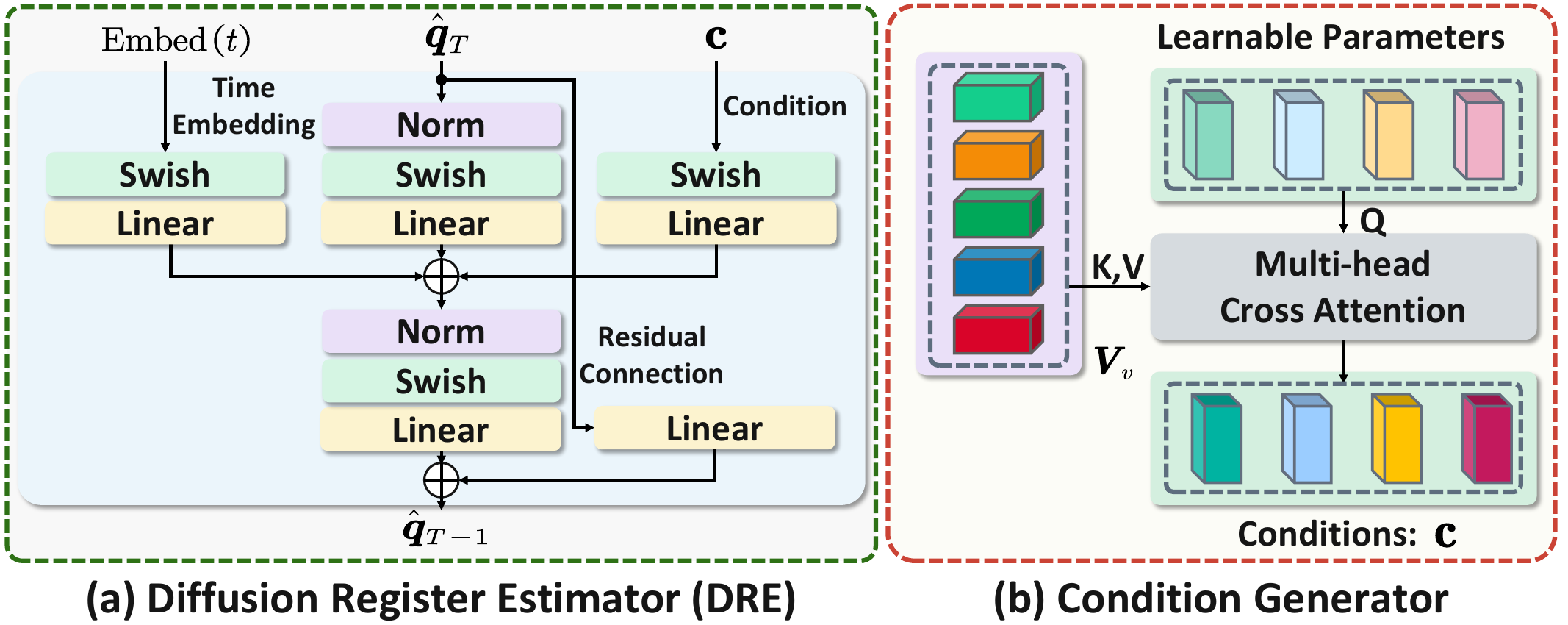}
    \end{subfigure}
    \caption{(a) Illustration of the proposed Diffusion Register Estimator Block (DRE). $\text{Embed}(t)$, $\bm{q}_{T}$ and $\bm{c}$ denote the temporal embedding, the latent embedding corrupted by $t$-step noise and the guided condition, respectively. (b) Condition generator for DRE.}
    \label{fig:appendix-arc}
\end{figure}

\noindent \textbf{Diffusion Register Estimator (DRE)} \quad As illustrated in \cref{fig:appendix-arc}(a), the DRE block follows an MLP-based architecture incorporating Layer Normalization \cite{ba2016layer}, activation functions, and linear projection layers. We use $N_{\text{dre}}=2$ blocks.

\noindent \textbf{Condition Generator} \quad The condition $\mathbf{c} \in \mathbb{R}^{N_r \times d}$ is obtained via a simple cross-attention mechanism between $\bmV_v \in \mathbb{R}^{N_v \times d}$ and learnable parameters, as illustrated in \cref{fig:appendix-arc} (b), and can be formulated as
\vspace{-.6em}
\begin{equation}
    \mathbf{c} = \text{CA}(LP, \bmV_v, \bmV_v) \in \mathbb{R}^{N_r \times d},
    \vspace{-.6em}
\end{equation}
where $\text{CA}$ denotes cross-attention, and $LP \in \mathbb{R}^{N_r \times d}$ represents learnable parameters.

\noindent \textbf{Asymmetric Attention Mask} \quad We retain the Gaussian self-attention as in \cite{GMMFormerV2,HLFormer} and instead define two cross-attention patterns through a designed masking strategy, as illustrated in Fig.~2(d). Given the global registers $\bmr_0$ and video embeddings $\bmV_o$, the cross-attention configurations are defined as follows.  
For video embeddings:  
\vspace{-.6em}
\begin{equation}
    \text{Query} = \bmV_o, \quad
    \text{Key} = \text{Value} = \texttt{Concat}([\bmV_o, \bmr_0]).
    \vspace{-.6em}
\end{equation}
For global registers:  
\vspace{-.6em}
\begin{equation}
    \text{Query} = \bmr_0, \quad
    \text{Key} = \text{Value} = \bmV_o.
    \vspace{-.6em}
\end{equation}

\begin{algorithm}[t] 
\caption{Register Generation Process during Training}
\label{algorithm:generation-code}
\begin{algorithmic}[1] 
\STATE \textbf{ \color{Green} Input:} Video features $\bmV_v$, all textual features from the video $\bmq$, condition features $\bmc$, timesteps $T$, noise schedule $\{\beta_t\}_{t=1}^T$, diffusion register estimator (DRE) $\epsilon_\phi$, probabilistic variational sampler (PVS),  textual perturbation sampler (TPS)
\STATE \textbf{ \color{Green} Output:} Optimal Registers $\bmr_0$, diffusion loss $L_D$

\STATE Initialize loss accumulator $L_D \leftarrow 0$
\STATE Precompute $\alpha_t = 1 - \beta_t$ and $\bar{\alpha}_t = \prod_{s=1}^t \alpha_s$ for all $t$
\STATE \textbf{\color{Orange} Feature Encoding}
\STATE $p(\bmr_T|\bmV_v)$ $\leftarrow$ PVS($\bmV_v$) \hfill \(\triangleright\) \textcolor{Orchid}{Using Eq.~(8)}
\STATE Sample $\bmr_T$ $\leftarrow$  $p(\bmr_T|\bmV_v)$ $\sim \mathcal{N} (\bm\mu_v, 
                    \bm\sigma^2_v I)$
\STATE $p(\bm{\hat{q}}|\bmq)$ $\leftarrow$ TPS($\bmq$) \hfill \(\triangleright\) \textcolor{Orchid}{Using Eq.~(6)}
\STATE Sample $\bm{\hat{q}}$ $\leftarrow$  $p(\bm{\hat{q}}|\bmq)$ $\sim \mathcal{N} (\bm\mu_{\hat{q}},\bm\sigma^2_{\hat{q}} I)$
\STATE \textbf{\color{Orange} Forward Diffusion Process}
\STATE ${\bm{\hat{q}}}_0$ $\leftarrow$ ${\bm{\hat{q}}}$
\FOR{$t=1$ \TO $T$}
    \STATE Sample  $\epsilon \sim \mathcal{N}(0, \mathbf{I})$  
    \STATE  ${\bm{\hat{q}}}_t \leftarrow \sqrt{\bar{\alpha}_t} {\bm{\hat{q}}}_0 + \sqrt{1 - \bar{\alpha}_t} \epsilon$
     \(\triangleright\) \hfill \textcolor{Orchid}{Add noise via Eq.~(13)}
    \STATE  $\hat{\epsilon} \leftarrow \epsilon_\phi({\bm{\hat{q}}}_t, t, \bmc)$ \hfill \(\triangleright\) \textcolor{Orchid}{Predict noise}
    \STATE  $L_{\text{dre}} \leftarrow ||\epsilon - \hat{\epsilon}||^2$ \hfill \(\triangleright\) \textcolor{Orchid}{Calculate loss via Eq.~(16)}
    \STATE  $L_D \leftarrow L_D + L_{\text{dre}}$ \hfill \(\triangleright\) \textcolor{Orchid}{Accumulate loss}
\ENDFOR
\STATE \textbf{\color{Orange} Reverse Generation Process}
\STATE Sample ${\bm{\hat{q}}}_T \leftarrow \bmr_T$ \hfill \(\triangleright\) \textcolor{Orchid}{Start generation}
\FOR{$t=T$ \TO $1$} 
    \STATE Predict noise $\hat{\epsilon} \leftarrow \epsilon_\phi({\bm{\hat{q}}}_t, t, \bmc)$
    \IF{$t > 1$} 
        \STATE Sample $z \sim \mathcal{N}(0, \mathbf{I})$
        \STATE ${\bm{\hat{q}}}_{t-1} \leftarrow \frac{1}{\sqrt{\alpha_k}} \left( {\bm{\hat{q}}}_t - \frac{1-\alpha_t}{\sqrt{1-\bar{\alpha}_t}} \hat{\epsilon} \right) + \sqrt{\beta_t} z$ 
    \ELSE
        \STATE ${\bm{\hat{q}}}_{t-1} \leftarrow \frac{1}{\sqrt{\alpha_t}} \left( {\bm{\hat{q}}}_{t} - \frac{1-\alpha_t}{\sqrt{1-\bar{\alpha}_t}} \hat{\epsilon} \right)$
    \ENDIF
\ENDFOR
\STATE $\bm{r}_0 \leftarrow {\bm{\hat{q}}}_0$ 
\STATE \textbf{return} $\bm{r}_0, L_D$ \hfill \(\triangleright\) \textcolor{Orchid}{Return features and loss}
\end{algorithmic}
\end{algorithm}
\subsection{Learning Objectives}
\noindent \textbf{Standard Similarity Retrieval Loss $L_{\text{sim}}$} \quad  Following prior works \cite{ms-sl, GMMFORMER, HLFormer}, we employ the widely adopted triplet loss \cite{dong2021dual} $L^{\text{trip}}$ and InfoNCE loss \cite{miech2020end, zhang2021video} $L^{\text{nce}}$ for PRVR. A text–video pair is treated as positive if the video contains a moment relevant to the query; otherwise, it is considered negative.
Given a positive pair $(Q, V)$, the triplet ranking loss over a mini-batch $\mathcal{B}$ is defined as follows:
 \vspace{-.6em}
\begin{gather}
L^{trip} = \frac{1}{n} \sum_{(Q,V) \in \mathcal{B}} \{\text{max}(0, m + S(Q^-, V) - S(Q, V)) \nonumber \\ 
    + \text{max}(0, m + S(Q, V^-) - S(Q,V))\},
     \vspace{-.6em}
\end{gather}
where $m$ denotes the margin, $Q^-$ and $V^-$ represent the negative text for $V$ and the negative video for $Q$, respectively, and the similarity score $S(\cdot,\cdot)$ is computed as in Eq.~(20).
 The infoNCE loss  is computed as:
  \vspace{-.6em}
\begin{gather}
L^{nce} = -\frac{1}{n} \sum_{(Q,V) \in \mathcal{B}} \{\text{log}(\frac{S(Q,V)}{S(Q,V) + \sum\nolimits_{Q_i^{-} \in \mathcal{N}_Q }S(Q_i^-, V)}) \nonumber\\
+ \text{log}(\frac{S(Q,V)}{S(Q,V) + \sum\nolimits_{V_i^{-} \in \mathcal{N}_V }S(Q, V_i^-)}) \},
 \vspace{-.6em}
\end{gather}
where $ \mathcal{N}_Q $ and $ \mathcal{N}_V $ represent the negative texts and videos of $ V $ and $ Q $ within the mini-batch $ \mathcal{B} $, respectively.
Finally , $L_{\text{sim}}$ is defined as:
 \vspace{-.6em}
\begin{equation}
L_{\text{sim}} = L_{c}^{\text{trip}} + L_{f}^{\text{trip}} + \lambda_c L_{c}^{\text{nce}} + \lambda_f L_{f}^{\text{nce}},
\vspace{-.6em}
\end{equation}
where $f$ and $c$ denote the objectives of the frame-scale and clip-scale branches, respectively and $\lambda_f$ and $\lambda_c$ are the corresponding hyper-parameters.

\noindent \textbf{Query Diversity Loss $L_{\text{div}}$} \quad Following \citet{GMMFormerV2}, given a collection of text queries in the mini-batch $\mathcal{B}$, the query diversity loss is defined as:
\vspace{-.6em}
\begin{equation}
    \begin{aligned}
&\!\!\!\!\ell(i,j) = (1 + \operatorname{cos}(\bmq_i, \bmq_j)) \operatorname{log}(1 + e^{\omega (\operatorname{cos}(\bmq_i, \bmq_j) + \delta)}),\\
&\!\!\!\!L_{\text{div}}= \frac{2}{M_q (M_q - 1)} \sum_{1 \leq i, j \leq M_q, i \neq j} \ell(i,j),        
\end{aligned}
\vspace{-.6em}
\end{equation}
where $\delta > 0$ is a margin factor,
$\omega > 0$ is a scaling factor and 
$M_q$ is the number of text queries relevant to a video. 

\subsection{Relationship between $L_{\text{DreamPRVR}}$ and $L_{\text{total}}$}
$L_{\text{DreamPRVR}}$ is the theoretical training objective defined in Eq.~(3). 
It consists of two components:  
(i) a KL-divergence term that enforces the registers to generate global contextual semantics consistent with the textual queries, and  
(ii) a likelihood term that strengthens video representation learning with register guidance, thereby facilitating improved cross-modal alignment and retrieval performance.

$L_{\text{total}}$ is the practical training objective, comprising four components: $L_{\text{tssl}}$, $L_{\text{pvs}}$, $L_{\text{dre}}$, and $L_{\text{sim}}$. 
Among them, $L_{\text{tssl}}$, $L_{\text{pvs}}$, and $L_{\text{dre}}$ jointly regularize the registers to generate text-consistent representations and capture richer textual semantics. These terms promote more effective register generation and correspond to optimizing the KL-divergence term in $L_{\text{DreamPRVR}}$. 
In addition, $L_{\text{sim}}$ serves as the retrieval-oriented similarity learning objective, aiming to improve retrieval performance. This term aligns with maximizing the likelihood component in $L_{\text{DreamPRVR}}$.

\subsection{Register Generation Process}
\noindent \textbf{Training Stage} \quad Please refer to  \cref{algorithm:generation-code}.

\noindent \textbf{Inference Stage} \quad The procedure follows \cref{algorithm:generation-code}, with the forward diffusion process and TPS sampling omitted.